\documentclass[runningheads]{llncs}

\usepackage{eccv}

\usepackage{eccvabbrv}

\usepackage{graphicx}
\usepackage{booktabs}
\usepackage{amsmath}
\usepackage{amssymb}
\usepackage{multirow}
\usepackage{booktabs}
\usepackage{color,colortbl}
\usepackage[dvipsnames]{xcolor}
\usepackage{pifont}
\usepackage{afterpage}
\usepackage{eso-pic}
\usepackage{indentfirst}
\usepackage{threeparttable}
\usepackage{caption}
\usepackage{wrapfig}
\usepackage{tikz}
\usepackage{textgreek}
\usepackage[accsupp]{axessibility}  

\definecolor{cvprblue}{rgb}{0.21,0.49,0.74}
\definecolor{Gray}{gray}{0.92}
\definecolor{hydra_attention_color}{RGB}{0,154,85}
\definecolor{ubin_red}{rgb}{0.8, 0, 0}
\definecolor{bluelike}{rgb}{0.05, 0.41, 0.95}
\definecolor{ao(english)}{rgb}{0.0, 0.5, 0.0}
\definecolor{indigo(dye)}{rgb}{0.0, 0.25, 0.42}
\definecolor{lasallegreen}{rgb}{0.03, 0.47, 0.19}
\definecolor{darkcyan}{rgb}{0.0, 0.55, 0.55}
\definecolor{lightseagreen}{rgb}{0.13, 0.7, 0.7}
\definecolor{persiangreen}{rgb}{0.0, 0.65, 0.58}
\definecolor{candypink}{rgb}{0.89, 0.44, 0.48}
\definecolor{cerisepink}{rgb}{0.93, 0.23, 0.51}
\definecolor{cerise}{rgb}{0.87, 0.19, 0.39}

\newcolumntype{L}[1]{>{\raggedright\let\newline\\\arraybackslash\hspace{0pt}}m{#1}}
\newcolumntype{C}[1]{>{\centering\let\newline\\\arraybackslash\hspace{0pt}}m{#1}}
\newcolumntype{R}[1]{>{\raggedleft\let\newline\\\arraybackslash\hspace{0pt}}m{#1}}

\usepackage[hyperfootnotes=false]{hyperref}
\hypersetup{colorlinks, citecolor=eccvblue}

\usepackage{orcidlink}
\usepackage{footmisc} 
\usepackage{authblk}

\title{Embedding-Free Transformer with Inference Spatial Reduction for Efficient Semantic Segmentation} 

\titlerunning{Embedding-Free Transformer with Inference Spatial Reduction}

\author{Hyunwoo Yu\thanks{Equal Contribution}\inst{1}\orcidlink{0009-0009-4426-8272} \and
Yubin Cho\protect\footnotemark[1]\inst{1,2}\orcidlink{0009-0001-8604-5431} \and
Beoungwoo Kang\protect\footnotemark[1]\inst{1} \and
Seunghun Moon\protect\footnotemark[1]\inst{1} \and
Kyeongbo Kong\protect\footnotemark[1]\inst{3}\orcidlink{0000-0002-1135-7502} \and
Suk-Ju Kang\thanks{Corresponding Author}\inst{1}\orcidlink{0000-0002-4809-956X}
}

\authorrunning{H. Yu et al.}

\institute{Department of Electronics Engineering, Sogang University, South Korea \and
AI Lab, CTO Division, LG Electronics, South Korea \and
Department of Electrical \& Electronics Engineering, Pusan National University, South Korea\\
\email{\{hyunwoo137, dbqls1219, beoungwoo, moonsh97, sjkang\}@sogang.ac.kr}\\
\email{kbkong@pusan.ac.kr}}

\begin{document}

\maketitle

\begin{abstract}
  We present an Encoder-Decoder Attention Transformer, ED-AFormer, which consists of the Embedding-Free Transformer (EFT) encoder and the all-attention decoder leveraging our Embedding-Free Attention (EFA) structure. The proposed EFA is a novel global context modeling mechanism that focuses on functioning the global non-linearity, not the specific roles of the query, key and value. For the decoder, we explore the optimized structure for considering the globality, which can improve the semantic segmentation performance. In addition, we propose a novel Inference Spatial Reduction (ISR) method for the computational efficiency. Different from the previous spatial reduction attention methods, our ISR method further reduces the key-value resolution at the inference phase, which can mitigate the computation-performance trade-off gap for the efficient semantic segmentation. Our EDAFormer shows the state-of-the-art performance with the efficient computation compared to the existing transformer-based semantic segmentation models in three public benchmarks, including ADE20K, Cityscapes and COCO-Stuff. Furthermore, our ISR method reduces the computational cost by up to 61\% with minimal mIoU performance degradation on Cityscapes dataset. The code is available at \url{https://github.com/hyunwoo137/EDAFormer}.
\end{abstract}

\section{Introduction}
\label{sec:intro}
Semantic segmentation, which aims to obtain the accurate pixel-wise prediction for the whole image, is one of the most fundamental tasks in the computer vision \cite{krizhevsky2012imagenet, long2015fully} and is widely used in various downstream applications \cite{cho2022class, cho2023cross,li2023semantic}. From the CNN-based models \cite{he2016deep, olaf2015unet, liu2022convnet, tan2021efficientnetv2, long2015fully, chen2017deeplab} to the transformer-based models \cite{vaswani2017attention, zheng2021rethinking, wang2021pyramid, wang2022pvt, liu2021swin, wu2021cvt, graham2021levit, fan2021mvit}, semantic segmentation models have been introduced in different structures. However, compared to other tasks, the semantic segmentation has a large amount of computation, as it treats the high resolution images and requires the per-pixel prediction decoder. Therefore, it is a significant challenge to explore the efficient structure for this task.

With the great success of the Vision Transformer \cite{dosovitskiy2020image} (ViT), recent semantic segmentation models \cite{xie2021segformer, guo2022segnext, yang2022lvt, ding2022davit, rao2021dynamicvit, chen2023sparsevit, yan2024multiscale, ranftl2021dpt, lu2023content, tang2023dynamic} mainly utilize the transformer-based structure to improve the performance by modeling the global context via the self-attention mechanism, and various advanced self-attention structures \cite{hatamizadeh2024fastervit,guo2022cmt, yuan2021t2t, zhang2021rest, chen2022mobileformer, li2022efficientformer, gong2021nasvit, liu2022nommer, sachin2021mobilevit, yang2022lvt, dong2022cswin, lin2023smt} have been introduced. In this paper, we analyze the general self-attention mechanism as two parts. The first is that the input feature is assigned the specific roles as the query, key and value by embedding the input features through the linear projection with the learnable parameters. The second is functioning as a global non-linearity, which obtains the attention weight between the query and the key via the softmax and then projects the attention weight into the value. We focus on that the real important part of global context modeling is the global non-linear functioning, not the specific roles (\textit{i.e.}, the query, key, and value) assigned to the input feature. We found that the simple but effective method, which removes the specific roles of the input feature, rather improves the performance. Therefore, we propose a novel self-attention structure, Embedding-Free Attention (EFA), which omits the embeddings of the query, key and value. 

With this powerful module, we also propose a semantic segmentation model, Encoder-Decoder Attention Transformer (EDAFormer), which is composed of the proposed Embedding-Free Transformer (EFT) encoder and the all-attention decoder. For the encoder, we adopt the hierarchical structure, and leverage our EFA module in the transformer blocks that effectively extract the global context features. For the decoder, inspired by \cite{zhang2022topformer,guo2022segnext, kang2024metaseg}, our all-attention decoder not only leverages our EFA, which effectively extracts the global context, but also is explored which level features need more global attention in the decoder. We empirically found that the higher level feature is more effective to consider the global context. Therefore, we design the all-attention decoder that leverages the more number of EFA modules to the higher level feature.

In addition, this paper addresses the issue of requiring additional training in the different structures whenever lighter (or less lightweight) models for lower computation (or higher accuracy). This issue causes user inconvenience and limits the versatility of lightweight methodologies. 

To solve this issue, we introduce a novel Inference Spatial Reduction (ISR) method that reduces the key-value resolution more at the inference phase than at the training phase. Our ISR exploits the Spatial Reduction Attention (SRA)-based structure in a completely different perspective with the existing SRA-based models \cite{wang2021pyramid, wu2021cvt, xie2021segformer, wang2022pvt, shim2023feedformer}, as we focus on making the reduction ratio different at training and inference. Through our method, the query learns a larger amount of the key and value information during training, and better copes with the reduced key and value during inference. This has the following two advantages. (1) Our method reduces the computational cost with little degradation in performance. (2) Our method allows to selectively adjust various computational costs of one pretrained model. 

We demonstrate the effectiveness of the proposed method in terms of the computational cost and performance on three public semantic segmentation benchmarks. Compared to the transformer-based semantic segmentation models, our model achieve the competitive performance in terms of the efficiency and the accuracy. Our contributions are summarized as follows:

\begin{itemize}
\item We propose a novel embedding-free attention structure that removes the specific roles of the query, key, and value but focuses on global non-linearity, thus achieving strong performance.
\item We introduce a semantic segmentation model, EDAFormer, which is designed with the EFT encoder and the all-attention decoder. Our decoder exploits the more number of the proposed EFA module at the higher level to capture the global context more effectively.
\item We propose a novel ISR method for the efficiency, which enables to reduce the computational cost with less degradation in performance at the inference phase and allows to selectively adjust the computational cost of the pretrained transformer model.
\item Our EDAFormer outperforms the existing transformer-based semantic segmentation models in terms of the efficiency and the accuracy on three public semantic segmentation benchmarks.
\end{itemize}

\section{Related Works}
\label{sec:relwork}
\subsection{Attention for Global Context}
The importance of modeling the global context has been demonstrated by the self-attention mechanism in the transformer. Beyond the general attention method, various attention methods have been studied.
\cite{wang2021pyramid, wang2022pvt} proposed the spatial reduction attention mechanism, which reduces the key-value resolution for efficiency. \cite{wu2022p2t} leveraged the pyramid pooling to reduce the key-value in multi-scale resolution. Based on the spatial reduction attention structure, \cite{guo2022cmt,zhang2021rest,zhang2021restv2} exploited the convolutional layer in the attention. The window-based attention method \cite{liu2021swin,liu2022swinv2} considered the local window regions for efficiency. \cite{chu2021twins} proposed the local window attention with global attention. The convolution-based attention \cite{wu2021cvt,xie2021segformer,guo2022cmt, dai2021coatnet} used the convolutional operation to consider local context with global context. The channel reduction attention method \cite{kang2024metaseg} reduced the query and key channels. However, all these self-attention methods are based on the query, key and value embeddings.
Different from these methods, we propose the efficient Embedding-Free Attention module by focusing on that the global non-linearity is important in the attention mechanism.

\subsection{Transformer-based Semantic Segmentation}
Since ViT \cite{dosovitskiy2020image} achieved the great performance in the image classification task, the transformer-based architectures have also been studied on the semantic segmentation, one of the most fundamental vision tasks. SETR \cite{zheng2021rethinking} was the first semantic segmentation model to adopt the transformer architecture as a backbone with convolutional decoder.
Beyond introducing the effective encoder structures, recent method \cite{xie2021segformer} proposed the efficient encoder-decoder structures for the semantic segmentation. SegFormer \cite{xie2021segformer} introduced a mix transformer encoder and a purely MLP-based decoder. FeedFormer \cite{shim2023feedformer} introduced a cross attention-based decoder to refer the low-level feature information of the transformer encoder. VWFormer \cite{yan2024multiscale} used the transformer encoder and exploited the window-based attention for considering the multi-scale representation in the decoder. We introduce the efficient Encoder-Decoder Attention TransFormer model for the semantic segmentation to effectively capture the global context at both the encoder and the decoder.

\section{Proposed Method}
\label{sec:method}
This section introduces our Encoder-Decoder Attention Transformer (EDAF-ormer), which is composed of the Embedding-Free Transformer (EFT) encoder and the all-attention decoder. Additionally, we describe our Inference Spatial Reduction (ISR) method that can reduce the computational cost effectively.

\begin{figure*}[t] 
\includegraphics[width=0.97\textwidth]{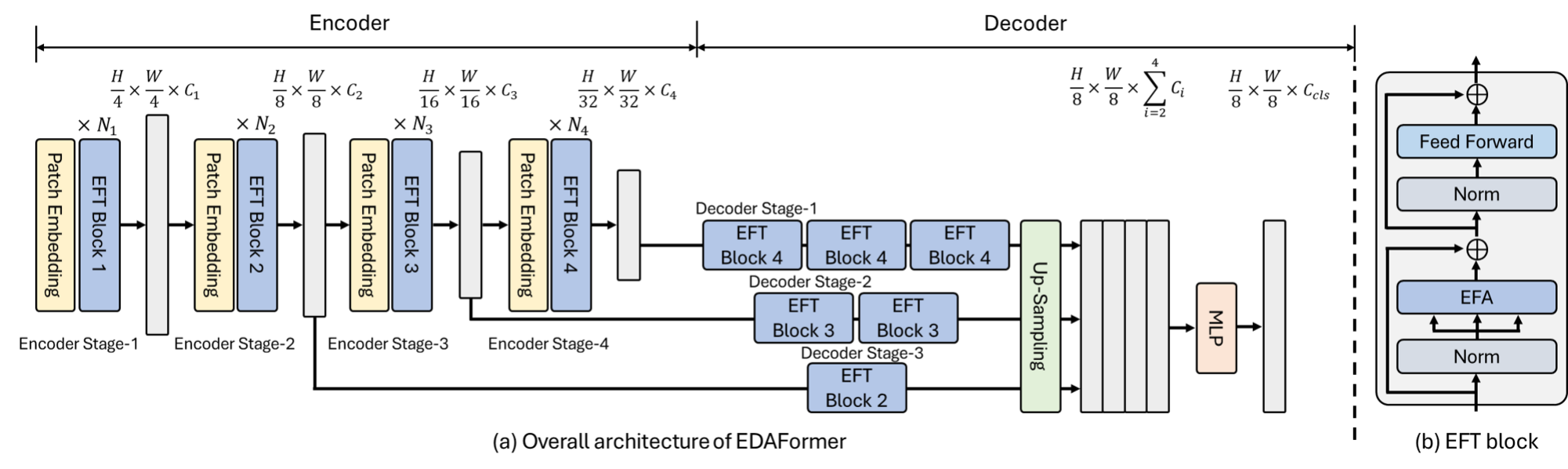}
\caption{(a) Overall architecture of the proposed EDAFormer, consisting of two main parts:  an EFT encoder and an all-attention decoder. The encoder and decoder of EDAFormer are designed with the query, key and value embedding free attention structure. (b) Details of the EFT block that contains EFA module.
} 
\label{intro_fig} 
\end{figure*}

\subsection{Overall Architecture}
\noindent\textbf{EDAFormer.}
As shown in \cref{intro_fig} (a), we leverage a hierarchical encoder structure, which is effective in the semantic segmentation task. When the input image is $ I \in \mathbb{R}^{H\times W\times3}$, the output feature of each stage is defined as $\textbf{F}_{i}\in\mathbb{R}^{\frac{H}{2^{i+1}} \times \frac{W}{2^{i+1}} \times C_i}$, where $i\in \{1,2,3,4\}$ denotes the index of the encoder stage, and $C$ is the channel dimension. At each stage, the features are first downsampled by the patch embedding block before being input to the transformer block. 

\begin{wrapfigure}{r}{0.43\textwidth} 
\centering 
\includegraphics[width=0.4\textwidth]{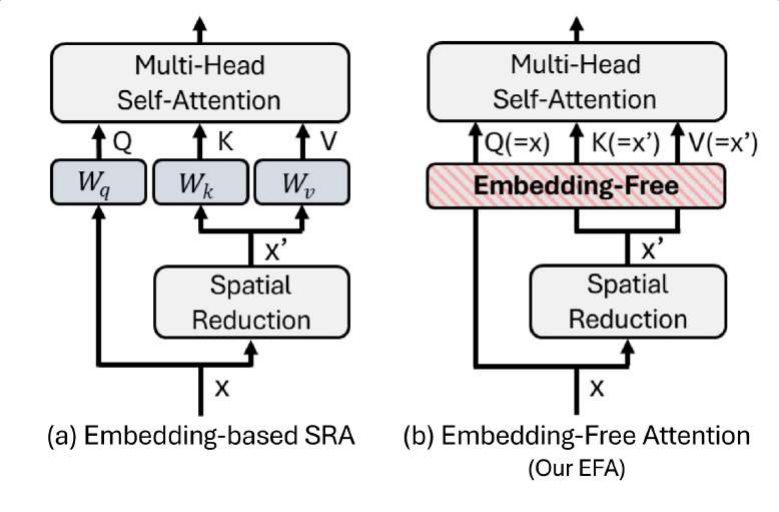}
\caption{Comparison of the previous method and our EFA.}
\label{fig:efa}
\end{wrapfigure}

As illustrated in \cref{intro_fig} (b), our transformer block structure of the encoder is composed of the Embedding-Free Attention (EFA) and the Feed-Forward Layer (FFL). As shown in \cref{fig:efa} (b), our EFA module omits the linear projection for the query \textbf{Q}, key \textbf{K} and value \textbf{V} embeddings, which are lightweight and effectively extracts the global context. Additionally, we adopt the spatial reduction attention (SRA) structure \cite{wang2022pvt} to leverage our ISR in the inference phase. We use the non-parametric operations and the average pooling to reduce the key-value spatial resolution, which has less impact on performance with the spatial reduction in the inference phase. The EFA module is formulated as follows: 
\begin{equation}
\begin{split}
    &\textbf{Q} = \textbf{x}_{in},\ \textbf{K}=\textbf{V} =\mathtt{SR}(\textbf{x}_{in},R), \\
    \textbf{Att} = \mathtt{so}&\mathtt{ftmax}(\textbf{Q}\cdot\textbf{K}^{T} /\sqrt{d_k}), \ \textbf{x}_{out}=\textbf{Att}\cdot\textbf{V},
\end{split}\end{equation}
where $\mathtt{SR}$ and $R$ denote the spatial reduction via the average pooling and the reduction ratio, respectively. $\textbf{x}_{in}$ is directly used as the query, and the spatial reduced features are used as the key-value. In the part where the softmax function is used for similarity scores between the query and the key, the global non-linearity can be applied to the input features, allowing the global context extraction without the specific roles of the query, key, and value. Then, the FFL is formulated as follows:
\begin{equation}
\mathtt{FFL}(\textbf{x}_{in}) = \mathtt{Linear}((\mathtt{DW}(\mathtt{Linear}(\textbf{x}_{in}))),
\end{equation}
\noindent
where $\mathtt{DW}$ indicates the depth-wise convolution. As the EFA and FFL are connected sequentially, the whole process of our EFT block is formulated as:
\begin{equation}
\begin{split}
&\textbf{z} =\mathtt{EFA}(\mathtt{LN}(\textbf{x}_{in})) + \textbf{x}_{in}, \\
&\textbf{x}_{out}=\mathtt{FFL}(\mathtt{LN}(\textbf{z}))+\textbf{z},
\end{split}\end{equation}
where $\textbf{z}$ is the intermediate features, and $\mathtt{LN}$ is a layer normalization. This embedding-free structure is effective for the classification and the semantic segmentation. In addition, we empirically find that our embedding-free structure is effective for our ISR in terms of considering the trade-off between the computation and the performance degradation.

\noindent\textbf{All-attention decoder.} As previous models \cite{yu2022vision,zhang2022topformer, guo2022segnext} have demonstrated, applying the SRA to the encoder features in the decoder is effective for capturing the global semantic-aware features. We thus design an all-attention decoder, which consists of  EFT blocks at all of the decoder stages. We also explore the optimal structure of the decoder for using EFT blocks. As a result, applying more attention blocks to the high-level features was effective for capturing globally more semantic informative features. As shown in \cref{intro_fig} (a), our decoder has a hierarchical structure that utilizes 3, 2, and 1 EFT blocks at the 1\(^{st}\) to 3\(^{rd}\) decoder stages, respectively. This structure is composed of a larger number of transformer blocks compared to the decoders of the previous transformer-based segmentation models, but has lower computational costs compared to previous models because the EFT block is lightweight.

In the all-attention decoder, the output features $\textbf{F}_i$ of each encoder stage $i \in \{2, 3, 4\}$ are first fed into the EFT blocks in each decoder stage $j\in \{3, 2, 1\}$, where $j$ denotes the index of the decoder stages. Then, the features $\widehat{\textbf{F}}_j\in\mathbb{R}^{H_j\times W_j\times C_j}$ of each decoder stage are up-sampled to $H_{2}\times W_{2}$ resolution using the bilinear interpolation. These up-sampled features $\textbf{U}_{j}\in\mathbb{R}^{H_2\times W_2\times C_j}$ are then concatenated and passed to linear layers for fusion. Finally, the final prediction mask is projected into the number of classes $C_{cls}$ mask by another linear layer. This process is formulated as:                      
\begin{equation}
\begin{split}
    &\widehat{\textbf{F}}_j = \mathtt{EFT}(\mathtt{LN}(\textbf{F}_{i})) + \textbf{F}_{i},\ \forall i\ \\
    {\textbf{U}_j}=\mathtt{Up}&\mathtt{sample}(\widehat{\textbf{F}}_j), \ \textbf{F}_{c} = \mathtt{Concat}({\textbf{U}_j}),\ \forall j \\
    &\textbf{M} = \mathtt{Linear}(\mathtt{Linear}(\textbf{F}_{c})),
\end{split}\end{equation}
where $\textbf{M} \in\mathbb{R}^{H_2\times W_2\times C_{cls}}$ is the final prediction mask.

\begin{figure*}[t] 
\includegraphics[width=\textwidth]{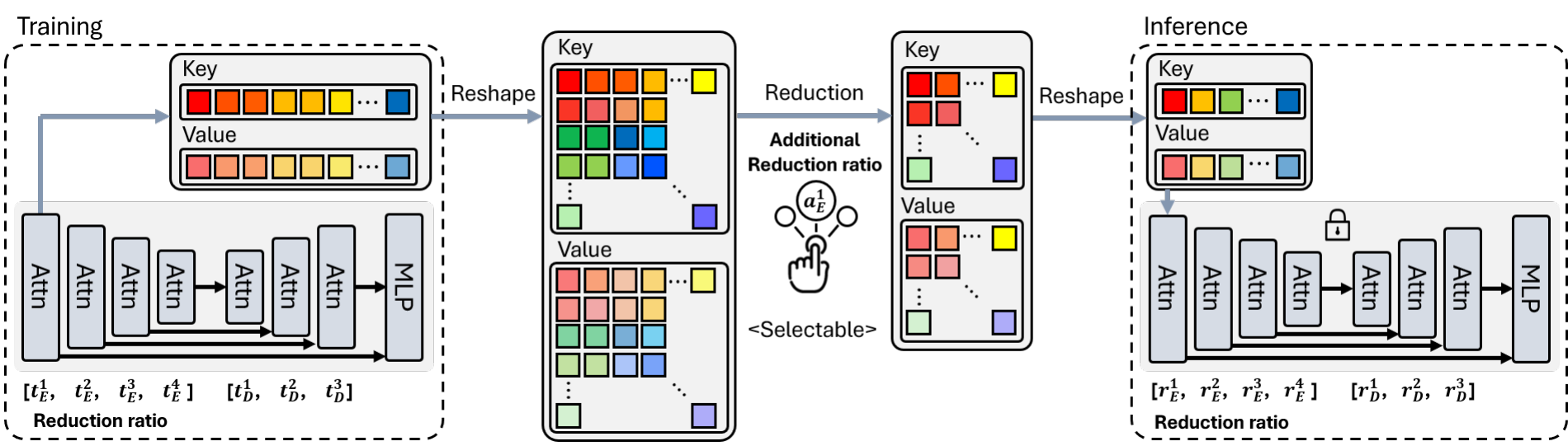}
\caption{Overview of our ISR method at the encoder stage-1. Our ISR applies the reduction ratio at the inference, reducing the key and value tokens selectively. This framework can be performed at every stage that contains the self-attention structure. It leads to flexibly reduce the computational cost without disrupting the spatial structure.
} 
\label{simpool} 
\end{figure*}

\subsection{Inference Spatial Reduction Method}
Different from previous SRA, our inference spatial reduction (ISR) method reduces the key-value spatial resolution at the inference phase. Our method achieves the computational efficiency by changing the hyperparameter associated with the `reduction ratio $R$' of the average pooling in the EFA module. Our ISR can be used in the self-attention structures because the self-attention has a special structure where reducing the resolution of key and value does not affect the shape of the input and output features. Due to this structure, the reduction ratio can be adjusted during inference without affecting the resolution of the input and output features. 

However, reducing the key and value resolution largely at training has the advantage of computational efficiency, but leads to the performance degradation because the query cannot consider enough information from the key and value. To address this issue, our ISR alleviates the trade-off gap between the computational cost and the accuracy by reducing the resolution of the key and value at inference. In this part, we describe that our ISR is applied to the our EDAFormer, which is the optimized architecture for applying our ISR effectively.

As shown in \cref{intro_fig}, our EDAformer uses the proposed transformer blocks in both encoder-decoder structures. Each pooling-based SRA used in each encoder stage and decoder stage has a corresponding reduction ratio setting that reduces the key and value resolution. At training as illustrated in \cref{simpool}, the reduction ratios $t_E^{i}$ of each encoder stage are set to \big[8, 4, 2, 1\big], which are the default setting of other previous models \cite{wang2021pyramid,xie2021segformer, wang2022pvt} using SRA. The reduction ratios $t_D^j$ of the decoder stage that takes each encoder features are set to \big[1, 2, 4\big], which are equal to the reduction ratios of the corresponding encoder stage. $t_E$ and $t_D$ denote the reduction ratio of the encoder and decoder at training, respectively.
The computational complexity of the previous attention is as follows:
\begin{equation}
    \mathrm{\Omega}(\mathtt{SRA}) = 2{\frac{(hw)^2}{r^2}}c,
\end{equation}
where $\mathrm{\Omega}$ and $\mathtt{SRA}$ denote the computational complexity and the spatial reduction attention. $h$, $w$ and $c$ represent the height, the width and the channel of the features, respectively. $r$ is the reduction ratio at training phase.

Under these reduction ratio settings, we train our EDAFormer to get pretrained weights. After that, at inference phase, it is possible to optionally adjust the inference computational reduction by selecting the reduction ratios at the discretion of the user. 
As shown in \cref{simpool}, $r_E^{i}$ and $r_D^{j}$ denote the reduction ratio of the encoder and decoder at inference, respectively. They are formulated as:
\begin{equation}
\begin{split}
    &r_E^i= t_E^i \times a_E^i,\ \forall i\\
    &r_D^j= t_D^j \times a_D^j,\ \forall j
\end{split}
\end{equation}
where $a_E^i$ and $a_D^j$ denote the additional reduction ratio of the encoder and decoder at inference, respectively. After applying our ISR, the computational complexity is as follows:
\begin{equation}
    \mathrm{\Omega}(\mathtt{ISR}(\mathtt{SRA})) = 2{\frac{(hw)^2}{r^2a^2}}c,
\end{equation}
where $\mathtt{ISR}$ is the inference spatial reduction and $a$ is the additional reduction ratio at inference. Therefore, one of the advantage of our ISR is that it is simple to obtain the computational reduction on the pretrained model without additional training. Our ISR reduces the performance degradation compared with reducing by $r^2a^2$ at training. Empirically, the optimal setting is \big[16,8,2,1\big]-\big[2,4,8\big] in the encoder-decoder, which has the best reduction ratio of the performance degradation to the computational cost reduction.

\section{Experiment}
\label{sec:experiment}

\begin{table*}[t]
\centering
\renewcommand{\arraystretch}{}
\resizebox{\textwidth}{!}{
\small
\begin{tabular}{l|c|cc|cc|cc}
\toprule[1.2pt]

\multirow{2}{*}{{Method}}&\multirow{2}{*}{{Params (M)}}& \multicolumn{2}{c|}{ADE20K}& \multicolumn{2}{c|}{Cityscapes}& \multicolumn{2}{c}{COCO-Stuff} \\ 
 & & {{GFLOPs $\downarrow$}} & {{mIoU (\%) $\uparrow$}} & {{GFLOPs $\downarrow$}} & {{mIoU (\%) $\uparrow$}} & {{GFLOPs $\downarrow$}} & {{mIoU (\%) $\uparrow$}} \\

\midrule

Segformer-B0 \cite{xie2021segformer} &3.8 &8.4 &37.4 &125.5 &76.2  &8.4 &35.6 \\
FeedFormer \cite{shim2023feedformer} &4.5 &7.8 &39.2 &107.4 &77.9  &- &- \\
VWFormer-B0 \cite{yan2024multiscale} & 3.7 & 5.1 & 38.9 &- &77.2 &5.1 &36.2 \\

\midrule

\cellcolor{Gray}\textbf{EDAFormer-T} \, (w/o ISR) &\cellcolor{Gray}{4.9} &\cellcolor{Gray}5.6 &\cellcolor{Gray}42.3  &\cellcolor{Gray}151.7  &\cellcolor{Gray}78.7  &\cellcolor{Gray}5.6 &\cellcolor{Gray}40.3 \\
\cellcolor{Gray}\textbf{EDAFormer-T} \, (w/\;\,  ISR) &\cellcolor{Gray}{4.9} &\cellcolor{Gray}\textbf{4.7} &\cellcolor{Gray}\textbf{42.1}  &\cellcolor{Gray}\textbf{94.9}  &\cellcolor{Gray}\textbf{78.7}  &\cellcolor{Gray}\textbf{4.7} &\cellcolor{Gray}\textbf{40.3} \\

\midrule[1.2pt]

OCRNet \cite{yuan2020ocrnet} &70.5 &164.8 &45.6 &1296.8  &81.1 & - & - \\
Swin UperNet-T \cite{liu2021swin} &60.0 &236.0 &44.4 &- &- &- &- \\
ContrastiveSeg \cite{wang2021exploring}&58.0&-&-&-&79.2&-&-\\
SenFormer \cite{bousselham2021senformer} &144.0 &179.0 &46.0 &- &- &- &- \\
Segformer-B2 \cite{xie2021segformer} &27.5 &62.4 &46.5 &717.1 &81.0 &62.4 &44.6 \\
ProtoSeg \cite{zhou2022rethinking}&90.5&-&48.6&-&80.6&-&42.4\\
MaskFormer \cite{cheng2021maskformer} &42.0 &55.0 &46.7 &- &- &- &- \\
Mask2Former \cite{cheng2022mask2former} &47.0 &74.0 &47.7 &- &- &- &- \\
FeedFormer-B2 \cite{shim2023feedformer} &29.1 &42.7 &48.0 &522.7 &81.5 &- &- \\
VWFormer-B2 \cite{yan2024multiscale} & 27.4 & 38.5 & 48.1 & - & 81.7 & 38.5 & 45.2\\

\midrule

\cellcolor{Gray}{\textbf{EDAFormer-B} \, (w/o ISR) } &\cellcolor{Gray}{29.4} &\cellcolor{Gray}32.0 &\cellcolor{Gray}49.0 &\cellcolor{Gray}605.9 &\cellcolor{Gray}81.6 &\cellcolor{Gray}32.0 &\cellcolor{Gray}45.9 \\
\cellcolor{Gray}{\textbf{EDAFormer-B} \, (w/\;\,  ISR) } &\cellcolor{Gray}{29.4} &\cellcolor{Gray}\textbf{29.4} &\cellcolor{Gray}\textbf{48.9} &\cellcolor{Gray}\textbf{452.9} &\cellcolor{Gray}\textbf{81.6} &\cellcolor{Gray}\textbf{29.4} &\cellcolor{Gray}\textbf{45.8} \\

\bottomrule[1.2pt]

\end{tabular}
}
\caption{Comparison with the transformer-based state-of-the-art semantic segmentation model on three public datasets. GFLOPs are computed using $512 \times 512$ resolutions for ADE20K and COCO-Stuff, and $2048 \times 1024$ resolutions for Cityscapes.}

\label{tab:table1}
\end{table*}

\subsection{Experimental Settings}
\noindent
\textbf{Datasets.} ADE20K \cite{zhou2017scene} is a challenging scene parsing dataset captured at indoors and outdoors. It consists of 150 semantic categories, and 20,210/2,000/3,352 images for training, validation, and testing. Cityscapes \cite{cordts2016cityscapes} is an urban driving scene dataset that contains 5,000 fine-annotated images with 19 semantic categories. It consists of 2,975/500/1,525 images in training, validation, and test sets. COCO-Stuff \cite{caesar2018coco} is a challenging dataset, which contains 164,062 images labeled with 172 semantic categories.

\noindent
\textbf{Implementation details.} The mmsegmentation codebase was used to train our model on 4 RTX 3090 GPUs. We pretrained our encoder on ImageNet-1K \cite{deng2009imagenet}, and our decoder was randomly initialized. For classification and segmentation evaluation, we adopted Top-1 accuracy and mean Intersection over Union (mIoU), respectively. We applied the same training settings and data augmentation as PVTv2\cite{wang2021pyramid} for ImageNet pretraining. We applied random horizontal flipping, random scaling with a ratio of 0.5-2.0 and random cropping with the size of 512$\times$512, 1024$\times$1024, and 512$\times$512 for ADE20K, Cityscapes, and COCO-Stuff, respectively. The batch size was 16 for ADE20K and COCO-Stuff, and 8 for Cityscapes. We used the AdamW optimizer for 160K iterations on ADE20K, Cityscapes and COCO-Stuff. 

\subsection{Comparison with State-of-the-art Methods}
\begin{wraptable}{r}{0.45\textwidth} 
\renewcommand{\arraystretch}{0.9} 
\resizebox{0.45\textwidth}{!}{
\small
\begin{tabular}{l|C{1.8cm}|C{1.5cm}|C{2.3cm}}
\toprule[1.4pt]
Models&{Params (M)}&{GFLOPs} & {Top-1 Acc. (\%)} \\ 
\midrule
RSB-ResNet-18 \cite{he2016deep,wightman2021resnet}  &12&1.8&70.6\\
PVTv2-B0 \cite{wang2022pvt}&3.4&0.6&70.5\\
MiT-B0 \cite{xie2021segformer}&3.7&0.6&70.5\\
\midrule
\cellcolor{Gray}\textbf{EFT-T (Ours)}&\cellcolor{Gray}3.7&\cellcolor{Gray}0.6&\cellcolor{Gray}\textbf{72.3}\\
\midrule[1.4pt]
ResNet50 \cite{he2016deep} & 25.5 & 4.1 & 78.5 \\
RSB-ResNet-152 \cite{he2016deep,wightman2021resnet} & 60.0 & 11.6 & 81.8 \\
DeiT-S \cite{touvron2021training} & 22.0 & 4.6 & 79.8 \\
PVT-Small \cite{wang2021pyramid} & 25.0 & 3.8 & 79.8 \\
PVTv2-B2 \cite{wang2022pvt} & 25.4 & 4.0 & 82.0 \\ 
MiT-B2 \cite{xie2021segformer} & 25.4 & 4.0 & 81.6 \\
T2T-ViT-14 \cite{yuan2021tokens} & 21.5 & 4.8 & 81.5 \\ 
TNT-S \cite{han2021tnt} & 23.8 & 4.8 & 81.5 \\
ResMLP-S24 \cite{touvron2022resmlp} & 30.0 & 6.0 & 79.4 \\ 
Swin-Mixer-T/D6 \cite{liu2021swin} & 23.0 & 4.0 & 79.7 \\
Visformer-S \cite{chen2021visformer} & 40.2 & 4.8 & 82.1 \\
gMLP-S \cite{liu2021pay} & 20.0 & 4.5 & 79.6 \\
PoolFormer-S36 \cite{yu2022metaformer}  & 31.0 & 5.0 & 81.4 \\ 
EfficientFormer-L3 \cite{li2022efficientformer} & 31.3 & 3.9 & 82.4 \\ 
FasterViT-0 \cite{hatamizadeh2024fastervit} & 31.4 & 3.3 & 82.1 \\
\midrule
\cellcolor{Gray}\textbf{EFT-B (Ours)}&\cellcolor{Gray}25.4&\cellcolor{Gray}4.2&\cellcolor{Gray}\textbf{82.4}\\
\bottomrule[1.4pt]
\end{tabular}}
\caption{Comparison with the previous models on ImageNet. GFLOPs were computed with 224$\times$224.}\label{tab:table22}
\end{wraptable}
\noindent
\textbf{Semantic segmentation.} In Table \ref{tab:table1}, we compared our EDAFormer with the previous transformer-based methods on three public datasets. The comparison includes the parameter size, FLOPs, and mIoU performance. Our lightweight model, EDAFormer-T (w/ ISR), showed 42.1\%, 78.7\% and 40.3\% mIoU, and our larger model, EDAFormer-B (w/ ISR), yielded 48.9\%, 81.6\% and 45.8\% mIoU on each dataset.
Compared to previous methods, both of our EDAFormer achieved the state-of-the-art performance with the efficient computation.

\noindent
\textbf{EFT encoder on ImageNet.} In \Cref{tab:table22}, we compared our Embedding-Free Transformer (EFT) encoder with the existing models on ImageNet-1K classification. Our EFT achieved higher performance than other transformer models. This result indicates that our EFT backbone is effective in the classification task by considering the spatial information globally even without the embeddings of the query, key and value.

\subsection{Effectiveness of our EFA at Decoder} 
To verify the effectiveness of considering the globality at the decoder, we compared the different operations at the Embedding-Free Attention (EFA) position of the EFT block in \Cref{tab:table2} (a). The applied operations are the local context operation (\textit{i.e.}, DW Conv, Conv) and the global context operation (\textit{i.e.}, w/ embedding attention, w/o embedding attention). Our w/o embedding structure improved 1.6\% and 2.4\% mIoU compared to the depth-wise convolution and the standard convolution, respectively. These results show that capturing the global context in the decoder is important for the mIoU performance improvement. While w/ embedding method outperformed the local context operation by capturing global context, our EFA further improved mIoU by 0.8\% with the lightweight model parameter and FLOPs. This indicates that our EFA module better models the global context.

\subsection{Structural Analysis of our All-attention Decoder} 
Our decoder, a \{3-2-1\} structure, is the hierarchical structure with six EFT blocks that assigns more attention blocks to high-level semantic features. In \Cref{tab:table2} (b), we verified the effectiveness of our decoder structure compared with three cases. The case of \{2-2-2\} structure assigned two EFT blocks equally to all decoder stages. The cases of \{1-2-3\}, \{1-4-1\} and our \{3-2-1\} allocated more EFT blocks to the decoder stage-3, 2 and 1, respectively. As a result, our \{3-2-1\} structure assigning more attention to higher level features shows better performance of 0.8\%, 1.7\%, 1.8\% mIoU compared to \{2-2-2\},\{1-2-3\}, and \{1-4-1\}, respectively. These results indicate that allocating the additional attention layers to the higher level features, which contain richer semantic information, is more effective for semantic segmentation performance.

\begin{table}[t]
\renewcommand{\arraystretch}{0.8}
\resizebox{\textwidth}{!}{
\small
\begin{tabular}{L{3.2cm}|C{2cm}|C{1.6cm}C{1.6cm}}
\toprule[1.1pt]

\multicolumn{4}{l}{(a) Effectiveness of our EFA for the decoder}\\
\midrule
\multirow{2}{*}{{Operation}} &\multirow{2}{*}{{Params (M)}}& \multicolumn{2}{c}{ADE20K} \\ 
& & \multicolumn{1}{c}{{GFLOPs}} & {{mIoU(\%)}} \\
\midrule
DW Conv &4.5 &5.1 &40.7 \\
Conv &6.6 &6.0 &39.9 \\
w/ embedding &5.7 &6.1 &41.5 \\
\cellcolor{Gray}\textbf{w/o embedding } &\cellcolor{Gray}{4.9} &\cellcolor{Gray}\textbf{5.6} & \cellcolor{Gray}\textbf{42.3} \\
\bottomrule[1.1pt]
\end{tabular}%

\hfil \quad

\resizebox{0.73\textwidth}{!}{
\small
\begin{tabular}{C{1.2cm}C{1.2cm}C{1.2cm}|c|C{1.5cm}C{1.5cm}}
\toprule[1.2pt]
\multicolumn{6}{l}{(b) Ablation on the number of EFA at each decoder stage}\\

\midrule
\multirow{2}{*}{{Stage-1}} &\multirow{2}{*}{{Stage-2}} &\multirow{2}{*}{{Stage-3}} &\multirow{2}{*}{{Params (M)}}& \multicolumn{2}{c}{ADE20K} \\ 
& & & & \multicolumn{1}{c}{{GFLOPs}} & {{mIoU(\%)}} \\
\midrule
2&2&2&4.6&5.7&41.5\\
1&2&3&4.2 &5.8 &40.6 \\
1&4&1&4.4&5.7&40.5\\
\cellcolor{Gray}\textbf{3}&\cellcolor{Gray}\textbf{2}&\cellcolor{Gray}\textbf{1}&\cellcolor{Gray}4.9&\cellcolor{Gray}\textbf{5.6}&\cellcolor{Gray}\textbf{42.3}\\
\bottomrule[1.1pt]
\end{tabular}}
}
\caption{Ablation studies of our all-attention decoder structure on the validation set of ADE20K. Our EFT encoder is used as the backbone.}
\label{tab:table2}
\end{table}

\begin{table*}[t]
\centering
\renewcommand{\arraystretch}{1.1}
\resizebox{\textwidth}{!}{
\begin{tabular}{cc|c|cc|cc|cc}
\toprule[1.4pt]
\multicolumn{2}{c|}{\big[ $r_{E}^{1},r_{E}^{2},r_{E}^{3},r_{E}^{4}$ \big]-\big[ $r_{D}^{1},r_{D}^{2},r_{D}^{3}$ \big] Reduction ratio}& \multirow{2}{*}{Params (M)} & \multicolumn{2}{c|}{{ADE20K}} & \multicolumn{2}{c|}{{Cityscapes}} & \multicolumn{2}{c}{{COCO-Stuff}} \\  
Train&Inference& &{{GFLOPs $\downarrow$}} & {{mIoU (\%) $\uparrow$} } & {{GFLOPs $\downarrow$}} & {{mIoU (\%) $\uparrow$} } & {{GFLOPs $\downarrow$}} & {mIoU (\%) $\uparrow$ } \\ 
\midrule
\multicolumn{9}{l}{(a) EDAFormer-T with the different reduction ratio at inference.}\\
\midrule
\multirow{4}{*}{\big[ 8, 4, 2, 1 \big]-\big[ 1, 2, 4 \big]}&\;\big[ 8, 4, 2, 1 \big]-\big[ 1, 2, 4 \big]$^{\dagger}$&{4.9}&5.6&42.3&151.7&78.7&5.6&40.3\\

&\big[ 8, 4, 2, 1 \big]-\big[ 2, 4, 8 \big]&{4.9} & {5.3 {\textcolor{hydra_attention_color}{(-5.4\%)}}}&{42.2  {\textcolor{ubin_red}{(-0.1)}}}&{133.6 {\textcolor{hydra_attention_color}{(-11.9\%)}}}&{78.7 {\textcolor{ubin_red}{(-0.0)}}}&{5.3 {\textcolor{hydra_attention_color}{(-5.4\%)}}}&{40.3 {\textcolor{ubin_red}{(-0.0)}}}\\

&\textbf{\big[16, 8, 2, 1\big]-\big[ 2, 4, 8 \big]}&{4.9} & {4.7 {\textcolor{hydra_attention_color}{(-16.1\%)}}}&{42.1 {\textcolor{ubin_red}{(-0.2)}}}&{94.9 {\textcolor{hydra_attention_color}{(-37.4\%)}}}&{78.7 {\textcolor{ubin_red}{(-0.0)}}}&{4.7 {\textcolor{hydra_attention_color}{(-16.1\%)}}}&{40.3 {\textcolor{ubin_red}{(-0.0)}}}\\

&\big[16, 8, 4, 2\big]-\big[ 2, 4, 8 \big]&{4.9} & {4.1 {\textcolor{hydra_attention_color}{(-26.8\%)}}}&{41.3 {\textcolor{ubin_red}{(-1.0)}}}&{59.1 {\textcolor{hydra_attention_color}{(-61.0\%)}}}&{78.1 {\textcolor{ubin_red}{(-0.6)}}}&{4.1 {\textcolor{hydra_attention_color}{(-26.8\%)}}}&{39.1 {\textcolor{ubin_red}{(-1.2)}}}\\

&\;\big[16, 8, 4, 2\big]-\big[ 2, 4, 8 \big]$^{*}$&{4.9} & {4.1 {\textcolor{hydra_attention_color}{(-26.8\%)}}}&{42.1 {\textcolor{ubin_red}{(-0.2)}}}&{59.1 {\textcolor{hydra_attention_color}{(-61.0\%)}}}&{78.6 {\textcolor{ubin_red}{(-0.1)}}}&{4.1 {\textcolor{hydra_attention_color}{(-26.8\%)}}}&{40.2 {\textcolor{ubin_red}{(-0.1)}}}\\

\midrule[1.4pt]
\multicolumn{9}{l}{(b) EDAFormer-B with the different reduction ratio at inference.}\\
\midrule
\multirow{4}{*}{\big[ 8, 4, 2, 1 \big]-\big[ 1, 2, 4 \big]}&\;\big[ 8, 4, 2, 1 \big]-\big[ 1, 2, 4 \big]$^{\dagger}$&{29.4}&32.0&49.0&605.9&81.6&32.0&45.9\\

&\big[ 8, 4, 2, 1 \big]-\big[ 2, 4, 8 \big]&{29.4} & {31.3 {\textcolor{hydra_attention_color}{(-2.2\%)}}}&{48.9 {\textcolor{ubin_red}{(-0.1)}}}&{569.0 {\textcolor{hydra_attention_color}{(-6.1\%)}}}&{81.6 {\textcolor{ubin_red}{(-0.0)}}}&{31.3 {\textcolor{hydra_attention_color}{(-2.2\%)}}}&{45.8 {\textcolor{ubin_red}{(-0.1)}}}\\

&\textbf{\big[16, 8, 2, 1\big]-\big[ 2, 4, 8 \big]}&{29.4} & {29.4 {\textcolor{hydra_attention_color}{(-8.1\%)}}}&{48.9 {\textcolor{ubin_red}{(-0.1)}}}&{452.9 {\textcolor{hydra_attention_color}{(-25.3\%)}}}&{81.6 {\textcolor{ubin_red}{(-0.0)}}}&{29.4 {\textcolor{hydra_attention_color}{(-8.1\%)}}}&{45.8 {\textcolor{ubin_red}{(-0.1)}}}\\

&\big[16, 8, 4, 2\big]-\big[ 2, 4, 8 \big]&{29.4} & {26.6 {\textcolor{hydra_attention_color}{(-16.9\%)}}}&{48.3 {\textcolor{ubin_red}{(-0.7)}}}&{298.1 {\textcolor{hydra_attention_color}{(-50.8\%)}}}&{81.4 {\textcolor{ubin_red}{(-0.2)}}}&{26.6 {\textcolor{hydra_attention_color}{(-16.9\%)}}}&{45.0 {\textcolor{ubin_red}{(-0.9)}}}\\

&\;\big[16, 8, 4, 2\big]-\big[ 2, 4, 8 \big]$^{*}$&{29.4} & {26.6 {\textcolor{hydra_attention_color}{(-16.9\%)}}}&{48.7 {\textcolor{ubin_red}{(-0.3)}}}&{298.1 {\textcolor{hydra_attention_color}{(-50.8\%)}}}&{81.6 {\textcolor{ubin_red}{(-0.0)}}}&{26.6 {\textcolor{hydra_attention_color}{(-16.9\%)}}}&{45.7 {\textcolor{ubin_red}{(-0.2)}}}\\

\bottomrule[1.4pt]
\end{tabular}
}
\caption{Computation and performance of our model on three standard benchmarks. $^{\dagger}$ indicates that the same reduction ratio is applied at training and inference. $^{\star}$ indicates the fine-tuning. \textbf{Bold} is optimal inference reduction ratio for our EDAFormer.}
\label{tab:table3}
\end{table*}

\subsection{Effectivness of our ISR in our EDAFormer} 
In \Cref{tab:table3}, we verified the effectiveness of our Inference Spatial Reduction (ISR) method in the proposed EDAFormer-T and EDAFormer-B, and empirically found the optimal reduction ratio. At training,
our EDAFormer was trained with the base setting of \big[8,4,2,1\big]-\big[1,2,4\big]. At inference, We experimented on applying our ISR to only decoder (\textit{i.e.}\big[8,4,2,1\big]-\big[2,4,8\big]), part of the encoder-decoder (\textit{i.e.}\big[16,8,2,1\big]-\big[2,4,8\big]), and all of the encoder-decoder (\textit{i.e.}\big[16,8,4,2\big]-\big[2,4,8\big]). The setting of \big[16,8,2,1\big]-\big[2,4,8\big] showed the optimal performance for improving the computational efficiency compared to the accuracy degradation. Compared to EDAFormer-T with the base setting, EDAFormer-T with the optimal setting reduced the computation by 16.1\%, 37.4\% and 16.1\% on ADE20K, Cityscapes and COCO-Stuff, respectively. 
The performance dropped by only 0.2\% mIoU on ADE20K and did not drop on COCO-Stuff and Cityscapes. Furthermore, EDAFormer-B reduced the computation by 8.1\% with only 0.1\% mIoU degradation on ADE20K and COCO-Stuff, and reduced the computation by 25.3\% without performance degradation on Cityscapes. These results indicate that our ISR method is simple, yet significantly reduces the computational cost with little performance degradation. In addition, our method showed the impressive effectiveness by only adjusting the reduction ratio at the inference without fine-tuning. Our ISR is effective without the fine-tuning, but we trained the models with 40K iterations for fine-tuning to further compensate for performance degradation at higher reduction ratio of [16,8,4,2]-[2,4,8]. As a result, EDAFormer-T showed a 0.2\% drop in mIoU on ADE20K, and 0.1\% drops in mIoU on Cityscapes and COCO-Stuff. EDAFormer-B showed 0.3\% and 0.2\% drops in mIoU on ADE20K and COCO-Stuff, and no drop in mIoU on Cityscapes.

\subsection{Comparison between the model with and without ISR.}
In Table \ref{tab:table5} (a), we compared our w/ ISR with w/o ISR, which used the same reduction ratio of \big[16,8,2,1\big]-\big[2,4,8\big] at both training and inference. 
Our EDAFormer with our ISR was trained with the reduction ratio of \big[8,4,2,1\big]-\big[1,2,4\big] and adjusted the ratio to \big[16,8,2,1\big]-\big[2,4,8\big] at inference. 
Despite the same computation at inference phase, the result with our ISR showed better mIoU than the case w/o ISR, with both 0.5\% improvements for our EDAFormer-T and EDAFormer-B, respectively. 
Therefore, our model w/ ISR, which considers enough information of the key and value during training, can achieve better performance than the model that cannot consider enough information by reducing the resolution of the key and value during training.

\begin{table}[t]
\renewcommand{\arraystretch}{1}
\resizebox{\textwidth}{!}{
\small
\resizebox{\textwidth}{!}{
\begin{tabular}{c|cc|cc}
\toprule[1.4pt]
\multirow{2}{*}{Method}&\multicolumn{2}{c|}{\big[ $r_{E}^{1},r_{E}^{2},r_{E}^{3},r_{E}^{4}$ \big]-\big[ $r_{D}^{1},r_{D}^{2},r_{D}^{3}$ \big] Reduction ratio}& \multicolumn{2}{c}{{COCO-Stuff}}\\
&Train&Inference&GFLOPs& {mIoU(\%)} \\
\midrule
\multicolumn{5}{l}{(a) Comparisons of our models with and without our ISR method}\\
\midrule
\multicolumn{5}{l}{EDAFormer-T}\\
\midrule

w/o ISR&{\big[16, 8, 2, 1\big]-\big[ 2, 4, 8 \big]}& \big[ 16, 8, 2, 1 \big]-\big[ 2, 4, 8 \big]&{4.7}&{39.8}\\

\cellcolor{Gray}w/ ISR& \cellcolor{Gray}\big[ 8, 4, 2, 1 \big]-\big[ 1, 2, 4 \big]&\cellcolor{Gray}\big[ 16, 8, 2, 1 \big]-\big[ 2, 4, 8 \big]&\cellcolor{Gray}{4.7}&\cellcolor{Gray}\textbf{40.3}\\

\midrule[1.4pt]
\multicolumn{5}{l}{EDAFormer-B}\\
\midrule

w/o ISR&{\big[16, 8, 2, 1\big]-\big[ 2, 4, 8 \big]}& \big[ 16, 8, 2, 1 \big]-\big[ 2, 4, 8 \big]&{29.4}&{45.3}\\

\cellcolor{Gray}w/ ISR& \cellcolor{Gray}\big[ 8, 4, 2, 1 \big]-\big[ 1, 2, 4 \big]&\cellcolor{Gray}\big[ 16, 8, 2, 1 \big]-\big[ 2, 4, 8 \big]&\cellcolor{Gray}{29.4}&\cellcolor{Gray}\textbf{45.8}\\

\bottomrule[1.4pt]
\end{tabular}%
}

\hfil \quad
\resizebox{\textwidth}{!}{
\small
\resizebox{0.98\textwidth}{!}
{\begin{tabular}{c|c|c|cc}
\toprule[1.4pt]
\multirow{2}{*}{Method} & Reduction ratio & \multirow{2}{*}{Params (M)} & \multicolumn{2}{c}{Cityscapes} \\

& \big[ $r_{E}^{1},r_{E}^{2},r_{E}^{3},r_{E}^{4}$ \big]-\big[ $r_{D}^{1},r_{D}^{2},r_{D}^{3}$ \big] & & GFLOPs & mIoU(\%) \\
\midrule
\multicolumn{5}{l}{{(b) Effectiveness of our EFA structure for our ISR}}\\
\midrule
\multirow{4}{*}{w/ embedding } & \big[ 8, 4, 2, 1 \big]-\big[ 1, 2, 4 \big] & 5.7 & 153.5 & 78.7 \\
 & \big[ 8, 4, 2, 1 \big]-\big[ 2, 4, 8 \big] & 5.7 & 134.7 & 78.5 {\textcolor{ubin_red}{(-0.2)}}\\
  & \big[ 8, 4, 2, 1 \big]-\big[ 3, 6, 9 \big] & 5.7 & 131.2 & 78.2 {\textcolor{ubin_red}{(-0.5)}}\\
   & \big[ 8, 4, 2, 1 \big]-\big[ 4, 8, 12 \big] & 5.7 & 130.0 & 77.9  {\textcolor{ubin_red}{(-0.8)}} \\

\midrule
\multirow{4}{*}{w/o embedding }& \big[ 8, 4, 2, 1 \big]-\big[ 1, 2, 4 \big] & 4.9& 151.7 &78.7 \\
 & \cellcolor{Gray}\big[ 8, 4, 2, 1 \big]-\big[ 2, 4, 8 \big] &\cellcolor{Gray}4.9 & \cellcolor{Gray}133.6 & \cellcolor{Gray}78.7 {\textcolor{ubin_red}{(-0.0)}}\\
  & \cellcolor{Gray}\big[ 8, 4, 2, 1 \big]-\big[ 3, 6, 9 \big] &\cellcolor{Gray}4.9 &\cellcolor{Gray}130.3 &\cellcolor{Gray}78.7 {\textcolor{ubin_red}{(-0.0)}}\\
   & \cellcolor{Gray}\big[ 8, 4, 2, 1 \big]-\big[ 4, 8, 12 \big] &\cellcolor{Gray}4.9 & \cellcolor{Gray}129.1 & \cellcolor{Gray}78.6 {\textcolor{ubin_red}{(-0.1)}} \\
   
\bottomrule[1.3pt]
\end{tabular}}
}}
\caption{(a) Ablation for mIoU (\%) performance comparisons of our models with and without our ISR method on COCO-Stuff. {(b)} Ablation for the effectiveness of our EFA structure for our ISR on Cityscapes \textit{val.}} 
\label{tab:table5}
\end{table}

\subsection{Effectiveness of Embedding-Free Structure for ISR} 
To verify the effectiveness of our embedding-free structure for ISR.  We experiment the ablated model that w/ embedding attention is adopt to our EFA 
position in all-attention decoder. We also compared with the ablated model (\textit{i.e.}, w/ embedding) by applying our ISR to the decoder stages in Table \ref{tab:table5} (b).
The w/ embedding structure showed the gradual performance degradation as the reduction ratio increased, and the reduction ratio of [8,4,2,1]-[4,8,12] showed the performance decrease of 0.8\% mIoU. However, our structure showed no performance degradation up to the reduction ratio of [8,4,2,1]-[3,6,9], and only a 0.1\% drop in mIoU at the reduction ratio of [8,4,2,1]-[4,8,12]. This indicate that our w/o embedding structure is effective with proposed ISR method.

\begin{table}[t]
  \begin{minipage}[t]{.5\linewidth}\centering
  \resizebox{\textwidth}{!}{\small
    \begin{tabular}{c|c|cc}
    \toprule[1.4pt]
    \multirow{2}{*}{Method} & Reduction ratio & \multicolumn{2}{c}{Cityscapes} \\
    & \big[ $r_{E}^{1},r_{E}^{2},r_{E}^{3},r_{E}^{4}$ \big]-\big[ $r_{D}^{1},r_{D}^{2},r_{D}^{3}$ \big] & mIoU (\%) $\uparrow$ & FPS (img/s) $\uparrow$ \\
    \midrule[1.2pt]
    \multicolumn{4}{l}{\textbf{(a)} Comparison of different spatial reduction methods for our ISR}\\
    \midrule
    w/o ISR & \big[ 8, 4, 2, 1 \big]-\big[ 1, 2, 4 \big] & 78.7 & 10.2 \\
    Bipartite matching \cite{bolya2022tome} & \big[ 11.2, 5.6, 2.8, 1.4 \big]-\big[ 1.4, 2.8, 5.6 \big] & 78.7 & 10.5  \\
    Max pooling & \big[ 16, 8, 2, 1 \big]-\big[ 2, 4, 8 \big] & 78.4 & 13.3  \\
    Overlapped pooling & \big[ 16, 8, 2, 1 \big]-\big[ 2, 4, 8 \big] & 78.7 & 13.2 \\
    Average pooling & \big[ 16, 8, 2, 1 \big]-\big[ 2, 4, 8 \big] & 78.7 & 13.3 \\
    \midrule[1.4pt]
    \multicolumn{4}{l}{\textbf{(b)} Inference speed improvement by increasing the reduction ratio}\\
    \midrule
    \multirow{4}{*}{Average pooling } & \big[ 8, 4, 2, 1 \big]-\big[ 1, 2, 4 \big] & 78.7 & 10.2 \\
     & \big[ 8, 4, 2, 1 \big]-\big[ 2, 4, 8 \big] & 78.7 & 11.0 \, \textcolor{bluelike}{(+7.8\%)}\\
      & \big[ 16, 8, 2, 1 \big]-\big[ 2, 4, 8 \big] & 78.7 & 13.2 \textcolor{bluelike}{(+29.4\%)}\\
       & \big[ 16, 8, 4, 2 \big]-\big[ 2, 4, 8 \big] & 78.1 & 15.0 \textcolor{bluelike}{(+47.1\%)}\\
    \bottomrule[1.4pt]
    \end{tabular}
    }\caption{{(a)} Performance and inference speed of our ISR with different spatial reduction methods. {(b)} Inference speed by increasing the reduction ratio.}\label{tab:table66}
  \end{minipage}%
  \hfill
  \begin{minipage}[t]{.45\linewidth}\centering
    \resizebox{\textwidth}{!}{\small
    \begin{tabular}{L{4.1cm}|C{2.5cm}|C{2.7cm}|C{2.7cm}}
    \toprule[1.4pt]
    Models&{Params (M)}&{GFLOPs $\downarrow$} & {mIoU (\%) $\uparrow$} \\ 
    \midrule
    CvT \cite{wu2021cvt}  &21.0&365.5&80.1\\
    \cellcolor{Gray}CvT + ISR  &\cellcolor{Gray}21.0&\cellcolor{Gray}222.6 {\textcolor{hydra_attention_color}{(-39.1\%)}}&\cellcolor{Gray}79.8 {\textcolor{ubin_red}{(-0.3)}}\\
    \midrule
    MViT \cite{fan2021mvit} &32.0&1435.6&80.5\\
    \cellcolor{Gray}MViT + ISR &\cellcolor{Gray}32.0&\cellcolor{Gray}838.0  {\textcolor{hydra_attention_color}{(-41.6\%)}}&\cellcolor{Gray}80.3 {\textcolor{ubin_red}{(-0.2)}}\\
    \midrule
    LVT \cite{yang2022lvt} & 5.0 & 132.1 & 79.6 \\
    \cellcolor{Gray}LVT + ISR & \cellcolor{Gray}5.0& \cellcolor{Gray}86.1  {\textcolor{hydra_attention_color}{(-34.8\%)}} &\cellcolor{Gray}79.5  {\textcolor{ubin_red}{(-0.1)}} \\
    \midrule
    Swin \cite{liu2021swin} &36.2&272.2&79.7\\
    \cellcolor{Gray}Swin + ISR &\cellcolor{Gray}36.2&\cellcolor{Gray}208.0  {\textcolor{hydra_attention_color}{(-23.6\%)}}&\cellcolor{Gray}79.0 {\textcolor{ubin_red}{(-0.7)}}\\
    \midrule
    DaViT \cite{ding2022davit} & 36.2 & 304.8 & 81.3 \\
    \cellcolor{Gray}DaViT + ISR &\cellcolor{Gray}36.2 &\cellcolor{Gray}242.0  {\textcolor{hydra_attention_color}{(-20.6\%)}}&\cellcolor{Gray}80.9 {\textcolor{ubin_red}{(-0.4)}} \\
    \midrule
    PVTv2 \cite{wang2022pvt} & 4.8 & 121.8 & 78.6 \\
    \cellcolor{Gray}PVTv2 + ISR &\cellcolor{Gray}4.8 &\cellcolor{Gray}63.4  {\textcolor{hydra_attention_color}{(-47.9\%)}} &\cellcolor{Gray}78.3 {\textcolor{ubin_red}{(-0.3)}} \\
    \midrule
    MiT \cite{xie2021segformer} &4.9 &117.4& 78.2\\
    \cellcolor{Gray}MiT \cite{xie2021segformer} + ISR &\cellcolor{Gray}4.9& \cellcolor{Gray}59.0 {\textcolor{hydra_attention_color}{(-49.7\%)}}& \cellcolor{Gray}77.6 {\textcolor{ubin_red}{(-0.6)}}\\
    \midrule
    SegFormer \cite{xie2021segformer} & 3.8 & 125.5 & 76.2 \\
    \cellcolor{Gray}SegFormer + ISR & \cellcolor{Gray}3.8 &\cellcolor{Gray}82.5  {\textcolor{hydra_attention_color}{(-34.3\%)}} & \cellcolor{Gray}75.6 {\textcolor{ubin_red}{(-0.6)}}\\
    \midrule
    FeedFormer \cite{shim2023feedformer} & 4.5 & 107.5& 77.9 \\ 
    \cellcolor{Gray}FeedFormer + ISR &\cellcolor{Gray}4.5 & \cellcolor{Gray}68.8  {\textcolor{hydra_attention_color}{(-36.0\%)}}&\cellcolor{Gray}77.4 {\textcolor{ubin_red}{(-0.5)}} \\ 
    \midrule
    EDAFormer (\textbf{Ours})&4.9 & 151.7&78.7\\
    \cellcolor{Gray}EDAFormer + ISR (\textbf{Ours})&\cellcolor{Gray}4.9 & \cellcolor{Gray}94.9  {\textcolor{hydra_attention_color}{(-37.4\%)}} &\cellcolor{Gray}78.7 {\textcolor{ubin_red}{(-0.0)}}\\
    \bottomrule[1.4pt]
    \end{tabular}
    }\caption{Applying our ISR without finetuning to various transformer-based models on Cityscapes \textit{val.}}\label{tab:table4}
  \end{minipage}
\end{table}

\subsection{Comparison of Spatial Reduction Methods for ISR}
In Table \ref{tab:table66} (a), We experimented to compare which method is better in terms of the mIoU and inference speed (FPS) for the key-value spatial reduction. The bipartite matching-based pooling had no mIoU degradation even though it was applied to every encoder-decoder stage. However, the bipartite matching can reduce maximum 50\% of tokens, which corresponds to a reduction ratio of $r=1.4$ ($\approx\sqrt{2}$). This is because it divides the tokens into two sets and merges them.  In addition, this method has the additional latency caused by the matching algorithm. Therefore, the bipartite matching showed similar FPS compared to w/o ISR even though they reduce the computation of the attention. The max pooling showed a drop of 0.3\% mIoU, and the overlapped pooling was slightly slower than the average pooling. Therefore, we adopted the average pooling method to reduce the tokens, which is a simple operation for general purposes and is most effective in terms of performance with inference speed.

\subsection{Inference Speed Enhancement}
In Table \ref{tab:table66} (b), we represented the inference speed (FPS) comparisons of various reduction ratios. We measured the inference speed by using a single RTX 3090 GPU without any additional accelerating techniques. Compared to base setting, applying our ISR shows 29.4\% and 47.1\% FPS improvements in the reduction ratios of [16,8,2,1]-[2,4,8] and [16,8,4,2]-[2,4,8], respectively. The inference speed became faster as the computational cost was reduced by increasing the reduction ratio. These results indicate that the the computational reduction by our ISR leads to the improvement of the actual inference speed. 

\subsection{Applying ISR to Various Transformer-based Models}
Our ISR can be universally applied not only to our EDAFormer, but also to other transformer-based models by using the additional spatial reduction at the inference. To verify generalizability of our ISR, we applied ours to various models in Table \ref{tab:table4}. The transformer-based backbones are trained with our decoder for the semantic segmentation task.
For the convolutional self-attention models (\textit{i.e.}, CvT \cite{wu2021cvt}, MViT \cite{fan2021mvit} and LVT \cite{yang2022lvt}), our ISR significantly reduced computation by 34.8$\sim$41.6\% with 0.1$\sim$0.3\% performance degradation. Our method also showed the effective computational reduction with less performance degradation for window-based attention models (\textit{i.e.}, Swin \cite{liu2021swin} and DaViT \cite{ding2022davit}), spatial reduction attention-based models (\textit{i.e.}, PVTv2 \cite{wang2022pvt} and MiT \cite{xie2021segformer}) and segmentation models (\textit{i.e.}, SegFormer \cite{xie2021segformer} and FeedFormer \cite{shim2023feedformer}). The result for FeedFormer using the cross-attention decoder showed that our method is also effective in the cross-attention mechanism. These results indicate that our ISR framework can be effectively extended to various transformer-based architecture using different attention methods, and our EDAFormer is especially the optimized architecture for applying our ISR effectively.

\begin{figure*}[t] 
\centering 
\includegraphics[width=\textwidth]{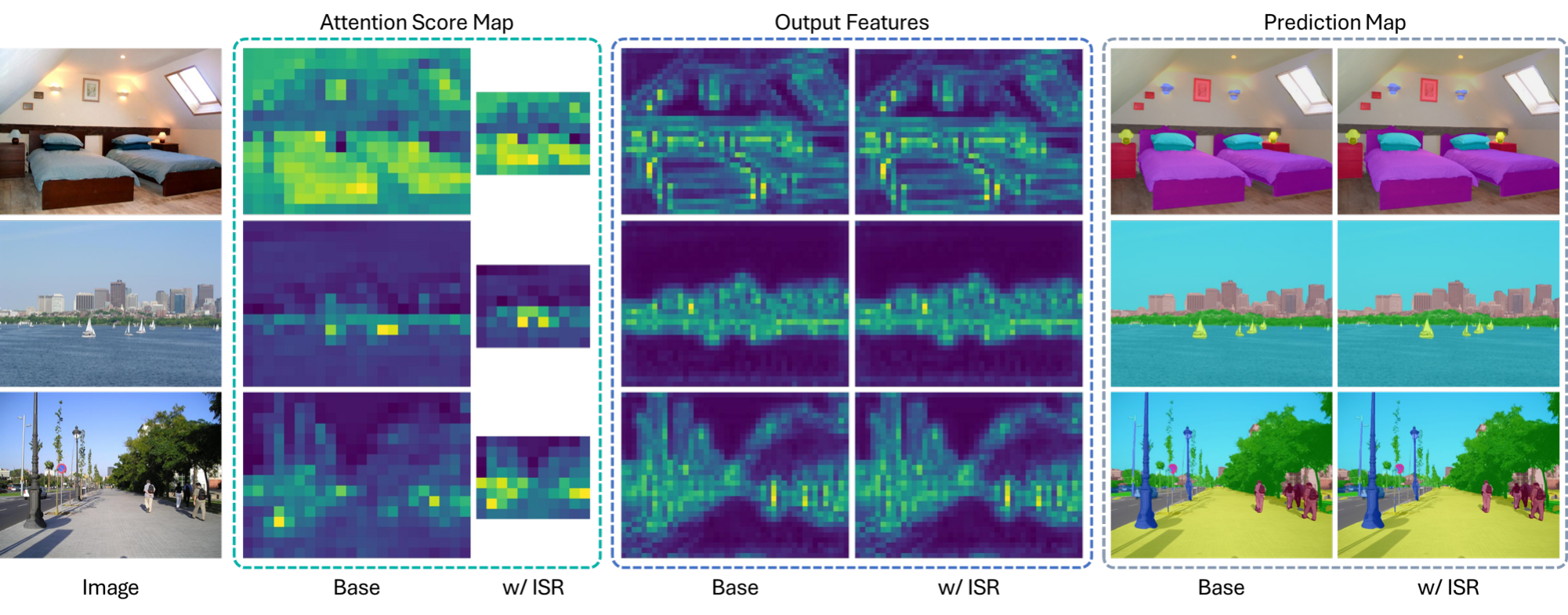} 
\caption{Visualization of the attention score map, output features, and prediction map on ADE20K. `Base' represents our EDAFormer trained with the base reduction ratio of [8,4,2,1]-[1,2,4]. `w/ ISR' represents our EDAFormer applied our ISR method.
} 
\label{feature} 
\end{figure*}

\subsection{Visualization of Features} 
\cref{feature} visualized the features and prediction maps of the EDAFormer-B decoder stage-2 before and after applying the ISR. Firstly, we visualized the attention score maps representing the similarity score between the query and key. When ISR was applied, the resolution of the attention score map was reduced because the resolution of the key was reduced. Compared to the similarity scores without applying the ISR, the similarity scores between the query and key applying the ISR were well maintained. In other words, the attention regions before and after applying ISR were similar, even though we reduce the key tokens rather than the attention score map. Therefore, this means that applying our ISR can maintain the semantic similarity scores in the global regions. 

Secondly, we compared the output features after operating between the attention score map and values. Surprisingly, the output features before and after applying ISR showed almost the same results. Therefore, these results indicate that the information obtained from the self-attention operation is maintained even though the spatial reduction is applied to the key and value in inference. Thirdly, when comparing the prediction maps, the results before and after applying the ISR are almost same. This means that the effect of ISR can be applied not only to the decoder stage-2, but also to the whole EDAFormer network.

\begin{figure*}[t] 
\centering 
\includegraphics[width=\textwidth]{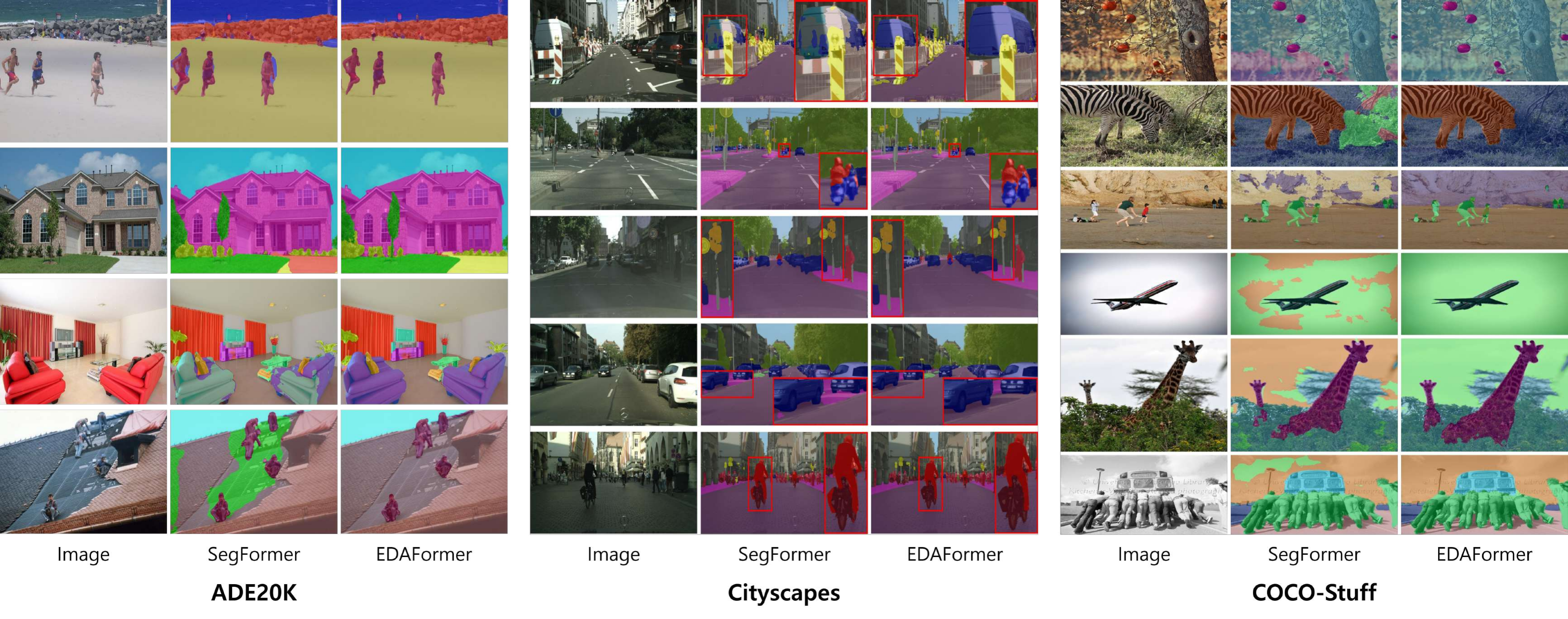} 
\caption{Qualitative results on ADE20K, Cityscapes, and COCO-Stuff. Compared to SegFormer, the predictions of our EDAFormer are more precise for various categories.} 
\label{quali} 
\end{figure*}

\subsection{Qualitative Results}
In \cref{quali}, we visualized our segmentation predictions on ADE20K, Cityscapes and COCO-Stuff, compared with the embedding-based transformer model (\textit{i.e.} SegFormer \cite{xie2021segformer}). Our EDAFormer better predicted the finer details near object boundaries. Our model also better segmented the large regions (\textit{e.g.}, road, roof and truck) than SegFormer. Furthermore, our model predicted the objects of the same category (\textit{e.g.}, sofa) that were far apart more precisely than SegFormer. This indicates that our embedding-free attention structure can capture enough global spatial information.

\section{Conclusion}
\label{sec:conclusion}
In this paper, we present an efficient transformer-based semantic segmentation model, EDAFormer, which leverages the proposed embedding-free attention module. The embedding-free attention structure can rethink the self-attention mechanism in the aspect of modeling the global context. In addition, we propose the novel inference spatial reduction framework for the efficiency, which changes the condition between train-inference phases. We hope that our attention mechanism and framework could further research efforts in exploring the lightweight and efficient transformer-based semantic segmentation model.

\section*{Acknowledgements}
This work was supported by Samsung Electronics Co., Ltd (IO201218-08232-01) and the National Research Foundation of Korea (NRF) grant funded by the Korea government(MSIT) (No. RS-2024-00414230) and MSIT (Ministry of Science and ICT), Korea, under the ITRC (Information Technology Research Center) support program (IITP-2024-RS-2023-00260091) supervised by the IITP (Institute for Information \& Communications Technology Planning \& Evaluation) and National Supercomputing Center with supercomputing resources including technical support(KSC-2023-CRE-0444).

%
%
\bibliographystyle{splncs04}
\bibliography{main}

\newpage
\appendix
\section*{Appendix}
\begin{itemize}
    \item In Appendix \ref{A}, we present performance-computation comparisons with our EDAFormer and other transformer-based state-of-the-art models.
    \item In Appendix \ref{attention}, we provide the computational analysis of our method in the attention block.
    \item In Appendix \ref{B}, we present fair comparisons of semantic segmentation decoder structures with the same backbone.
    \item In Appendix \ref{C}, we present the comparisons of computations and performance on various reduction ratios.
    \item In Appendix \ref{fps}, we present the FPS comparison with other segmentation models.
    \item In Appendix \ref{backbones}, we provide the in-depth analysis for the effectiveness of our embedding-free structure.
    \item In Appendix \ref{D}, we provide the additional visualizations of features before and after applying our ISR.
    \item In Appendix \ref{E}, we provide qualitative results compared with the proposed model and previous state-of-the-art models on ADE20K, Cityscapes and COCO-Stuff datasets.
\end{itemize}

\begin{minipage}{0.98\textwidth}\centering
  \begin{minipage}[b]{0.49\textwidth}
    \centering
    \renewcommand{\arraystretch}{}
        \resizebox{0.8\textwidth}{!}{
        \small
        \begin{tabular}{l|c|c|c}
        \toprule[1.2pt]
        Models&{Params (M)}&{GFLOPs $\downarrow$} & {mIoU (\%) $\uparrow$} \\ 
        \midrule
        PVT-Tiny \cite{wang2021pyramid}&15.5&32.2&32.9 \\
        SegFormer-B0 \cite{xie2021segformer}&3.8&8.4&37.4 \\
        RTFormer-Slim \cite{wang2022rtformer}&4.8&17.5&36.7 \\
        FeedFormer-B0 \cite{shim2023feedformer}&4.5&7.8&39.2 \\
        VWFormer-B0 \cite{yan2024multiscale}&3.7&5.1&38.9 \\
        \midrule
        \cellcolor{Gray}\textbf{EDAFormer-T} \, (w/o ISR) &\cellcolor{Gray}4.9&\cellcolor{Gray}5.6&\cellcolor{Gray}42.3\\
        \cellcolor{Gray}\textbf{EDAFormer-T} \, (w/\;\,  ISR) &\cellcolor{Gray}4.9&\cellcolor{Gray}\textbf{4.7}&\cellcolor{Gray}\textbf{42.1}\\
        \midrule[1.2pt]
        PVT-Medium \cite{wang2021pyramid}&48.0&61.0&41.6 \\ 
        Focal-T \cite{yang2021focal}&62.0&998.0&45.8 \\
        Twins-SVT-UperNet-S \cite{chu2021twins}&54.4&228.0&46.2 \\
        SegFormer-B2 \cite{xie2021segformer}&27.5&62.4&46.5 \\
        MaskFormer \cite{cheng2021maskformer}&42.0&55.0&46.7 \\
        SenFormer \cite{bousselham2021senformer}&144.0&179.0&46.0 \\
        CrossFormer-S \cite{wang2022crossformer}&62.3&968.5&47.4 \\
        MPViT-S \cite{lee2022mpvit}&52.0&943.0&48.3 \\
        DW-T \cite{ren2022dwvit}&61.0&953.0&45.7 \\
        MixFormer-B1 \cite{chen2022mixformer}&35.0&854.0&42.0 \\
        DAT-T \cite{xia2022dat}&32.0&198.0&42.6 \\
        NomMer-T \cite{liu2022nommer}&54.0&954.0&46.1 \\
        Shunted-S\cite{ren2022shunted}&52.0&940.0&48.9 \\
        Mask2Former \cite{cheng2022mask2former}&47.0&74.0&47.7 \\
        DaViT-Tiny \cite{ding2022davit}&60.0&940.0&46.3 \\
        Scalable ViT-S \cite{yang2022scalable}&57.0&931.0&48.5 \\
        Ortho-S \cite{huang2021ortho}&54.0&956.0&48.5 \\
        RTFormer-Base \cite{wang2022rtformer}&16.8&67.4&42.1 \\
        FeedFormer-B2 \cite{shim2023feedformer}&29.1&42.7&48.0 \\
        GC ViT-T \cite{hatamizadeh2023gcvit}&58.0&947.0&47.0 \\
        Flatten-Swin-T \cite{han2023flatten}&60.0&946.0&44.8 \\
        VWFormer-B2 \cite{yan2024multiscale}&27.4&38.5&48.1 \\
        \midrule
        \cellcolor{Gray}\textbf{EDAFormer-T} \, (w/o ISR) &\cellcolor{Gray}29.4&\cellcolor{Gray}32.0&\cellcolor{Gray}49.0\\
        \cellcolor{Gray}\textbf{EDAFormer-T} \, (w/\;\,  ISR) &\cellcolor{Gray}29.4&\cellcolor{Gray}\textbf{29.4}&\cellcolor{Gray}\textbf{48.9}\\
        \bottomrule[1.2pt]
      \end{tabular}}\vspace{-0.3cm}
      \captionsetup[table]{hypcap=false}
      \captionof{table}{Comparison with previous transformer-based models on ADE20K. GFLOPs were computed with 512$\times$512.}
    \end{minipage}\label{tab:tableseg}
    \hfill 
    \begin{minipage}[b]{0.49\textwidth}
    \centering
    \includegraphics[width=\textwidth]{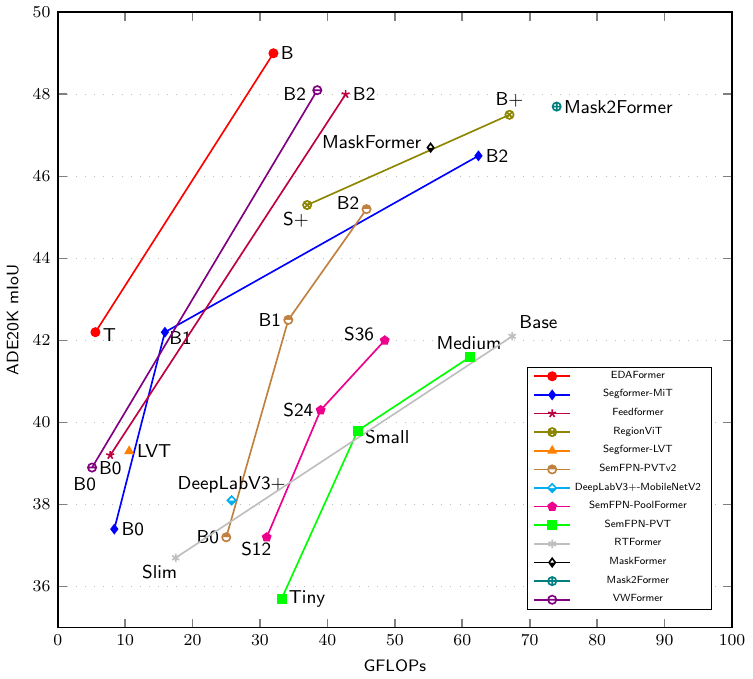}\vspace{-0.3cm}
    \captionsetup[figure]{hypcap=false}
    \captionof{figure}{Performance-Computation curves of our EDAFormer and existing segmentation models on ADE20K.}\label{fig:graph1_1}
    \end{minipage}
  \end{minipage}

\section{Additional Comparison to Transformer-based Models}
\label{A}
In Table \ref{tab:tableseg}, we compared with additional transformer-based models on ADE20K \cite{zhou2017scene} validation set as most transformer-based backbone studies include ADE20K results to show the semantic segmentation performance. We showed 4.7 GFLOPs with 42.1\% mIoU and 29.4 GFLOPs with 48.9\% mIoU in EDAFormer-T and EDAFormer-B (w/ ISR), respectively. In addition, Fig. \ref{fig:graph1_1} presented the performa-nce-computation curves, which include the comparisons with lightweight transfo-rmer-based models. These results showed that our EDAFormer achieved the efficient and significant performance compared to previous transformer-based state-of-the-art models.

\begin{table}[t]
\centering
\renewcommand{\arraystretch}{}
\begin{minipage}[t]{\textwidth} 
\centering
\resizebox{\textwidth}{!}{ 
\small
\begin{tabular}{c|cc|cc|cc|cc|cc}
\toprule
 \multirow{2}{*}{Mechanism}& \multicolumn{2}{c|}{QKV Embedding} & \multicolumn{2}{c|}{Global Functioning}& \multicolumn{2}{c|}{Output Projection} & \multicolumn{2}{c|}{Others} & \multicolumn{2}{c}{Total} \\
& {MFLOPs $\downarrow$} & {Params (K)} & {MFLOPs $\downarrow$} & {Params (K)}&{MFLOPs $\downarrow$} & {Params (K)} &{MFLOPs $\downarrow$} & {Params (K)}&{MFLOPs $\downarrow$} & {Params (K)}\\ 
\midrule
\textOmega\, (SRA) & 4.82 & 49.6 & 2.46 & 0.0& 3.21& 16.5& 0.83 & 16.5&11.32&82.6 \\
\cellcolor{Gray}\textOmega\, (EFA w/o ISR) & \cellcolor{Gray}0.00 & \cellcolor{Gray}0.0 &\cellcolor{Gray}2.46 &\cellcolor{Gray}0.0&\cellcolor{Gray}3.21&\cellcolor{Gray}16.5&\cellcolor{Gray}0.83 &\cellcolor{Gray}16.5&\cellcolor{Gray}\textbf{6.50} \textcolor{hydra_attention_color}{\textbf{(-42.6\%)}}&\cellcolor{Gray}\textbf{33.0} \textcolor{hydra_attention_color}{\textbf{(-60.0\%)}} \\
\cellcolor{Gray}\textOmega\, (EFA w/\,\,\, ISR) & \cellcolor{Gray}0.00 & \cellcolor{Gray}0.0 & \cellcolor{Gray}0.61 & \cellcolor{Gray}0.0&\cellcolor{Gray}3.21& \cellcolor{Gray}16.5& \cellcolor{Gray}0.18& \cellcolor{Gray}16.5&\cellcolor{Gray}\textbf{4.00} \textcolor{hydra_attention_color}{\textbf{(-64.7\%)}}&\cellcolor{Gray}\textbf{33.0} \textcolor{hydra_attention_color}{\textbf{(-60.0\%)}}\\
\bottomrule
\end{tabular}
}
\end{minipage}
\hspace{-0.3cm} 
\begin{minipage}[t]{\columnwidth} 
\centering
\includegraphics[width=\textwidth]{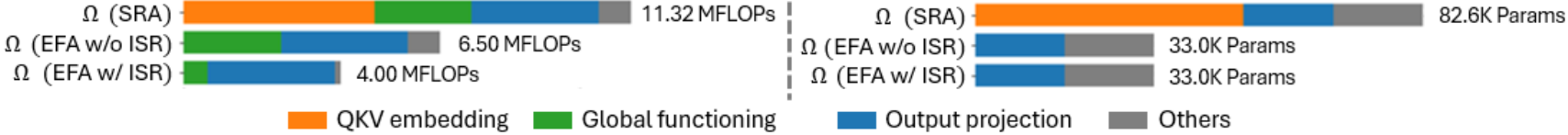}
\end{minipage}
\caption{Computational analysis of our method in the attention block. The FLOPs and parameters were computed on stage 3 features of 224 $\times$ 224 size.}
\label{tab:att}
\end{table}

\section{Computational Analysis in Attention Block}\label{attention}
In Table \ref{tab:att}, we compared the computation of the attention to analyze the effectiveness of our embedding-free structure and our inference spatial reduction (ISR) method. We analyzed the attention mechanism by dividing into the query-key-value embeddings, the global functioning, the output projection and others. Since our structure is based on spatial reduction attention (SRA), others are the spatial reduction operation. Our embedding-free structure effectively reduced the total MFLOPs by 42.6\% and the parameters by 60.0\%. In addition, our ISR reduced the computation of the global functioning. Therefore, compared to the original SRA, our embedding-free structure with ISR reduced the MFLOPs by 64.7\% and the parameters by 60.0\%.

\section{Comparison of Decoder Structure with Same Backbone}\label{B}

\begin{wraptable}{r}{0.5\textwidth}\vspace{-0.9cm}
\centering
\renewcommand{\arraystretch}{}
\resizebox{0.5\textwidth}{!}{
\small
\begin{tabular}{C{1.7cm}|c|c|c}
\toprule[1.2pt]
Encoder&Decoder&{GFLOPs $\downarrow$} & {mIoU (\%) $\uparrow$} \\
\midrule
\multirow{5}{*}{MIT-B0}&SegFormer \cite{xie2021segformer}&8.4&37.4 \\
&FeedFormer \cite{shim2023feedformer}&7.8&39.2 \\
&SegNeXt \cite{guo2022segnext}&5.2&38.7 \\
&VWFormer \cite{yan2024multiscale}&5.1&38.9 \\
&\cellcolor{Gray}\textbf{EDAFormer(Ours)}&\cellcolor{Gray}\textbf{4.6}&\cellcolor{Gray}\textbf{40.1}\\
\bottomrule[1.2pt]
\end{tabular}
}\vspace{-0.3cm}
\caption{\label{tab:decoder}Comparison of our EDAFormer and other segmentation methods with a MiT backbone \cite{xie2021segformer} on ADE20K. \vspace{-1cm}}
\end{wraptable}


As the backbone has a significant impact on the semantic segmentation performance, we experimented with other segmentation methods using the same backbone for a more fair comparison of the decoder structure in Table \ref{tab:decoder}. We use a Mix Transformer (MiT) structure as a common backbone, which is widely used as a transformer-based backbone in the semantic segmentation. In the decoder, we compared our EDAFormer (\textit{i.e.}, All-attention decoder) with previous powerful methods, including SegFormer \cite{xie2021segformer} (\textit{i.e.}, All-MLP decoder), FeedFormer \cite{shim2023feedformer} (\textit{i.e.}, Feature query decoder), SegNeXt \cite{guo2022segnext} (\textit{i.e.}, Ham decoder), VWFormer \cite{yan2024multiscale} (\textit{i.e.}, Multi-scale decoder). As shown in Table \ref{tab:decoder}, our EDAFormer showed the most efficient computational cost with remarkable mIoU performance by modeling the global context.

\begin{table}[t]
\renewcommand{\arraystretch}{} 
\centering
\resizebox{0.4\textwidth}{!}{
\begin{tabular}{c|cc}
\toprule[1.2pt]
\multirow{2}*[1.5ex]{Reduction ratio} & \multicolumn{2}{c}{{Cityscapes}} \\  
{\begin{normalsize}\big[ $r_{E}^{1},r_{E}^{2},r_{E}^{3},r_{E}^{4}$ \big]-\big[ $r_{D}^{1},r_{D}^{2},r_{D}^{3}$ \big]\end{normalsize}}  & {{GFLOPs $\downarrow$}} & {{mIoU (\%) $\uparrow$}} \\ 
\midrule[1.2pt]
\, \big[ 8 - 4 - 2 - 1 \big]-\big[ 1 - 2 - 4 \big]$^{\dagger}$  & 151.7 & 78.7 \\
\big[ 8 - 4 - 2 - 1 \big]-\big[ 1 - 2 - 8 \big]  & 145.3 & 78.7 \\
\big[ 8 - 4 - 2 - 1 \big]-\big[ 1 - 4 - 8 \big]  & 138.8 & 78.7 \\
\big[ 8 - 4 - 2 - 1 \big]-\big[ 2 - 4 - 8 \big]  & 133.6 & 78.7 \\
\midrule
\big[16 - 4 - 2 - 1 \big]-\big[ 1 - 2 - 4 \big] & 125.9 & 78.7 \\
\big[16 - 4 - 2 - 1 \big]-\big[ 1 - 2 - 8 \big] & 119.5 & 78.7 \\
\big[16 - 4 - 2 - 1 \big]-\big[ 1 - 4 - 8 \big] & 113.0 & 78.7 \\
\big[16 - 4 - 2 - 1 \big]-\big[ 2 - 4 - 8 \big] & 107.8 & 78.7 \\
\midrule
\big[16 - 8 - 2 - 1 \big]-\big[ 1 - 2 - 4 \big]  & 113.0 & 78.7 \\
\big[16 - 8 - 2 - 1 \big]-\big[ 1 - 2 - 8 \big]  & 106.6 & 78.7 \\
\big[16 - 8 - 2 - 1 \big]-\big[ 1 - 4 - 8 \big]  & 100.1 & 78.7 \\
\cellcolor{Gray}{{\big[16 - 8 - 2 - 1 \big]-\big[ 2 - 4 - 8 \big]}}  & \cellcolor{Gray}{94.9} & \cellcolor{Gray}{78.7}  \\
\midrule
\big[16 - 8 - 4 - 1 \big]-\big[ 1 - 2 - 4 \big]  & 80.6 & 78.1 \\
\big[16 - 8 - 4 - 1 \big]-\big[ 1 - 2 - 8 \big]  & 74.1 & 78.1 \\
\big[16 - 8 - 4 - 1 \big]-\big[ 1 - 4 - 8 \big]  & 67.6 & 78.1 \\
\big[16 - 8 - 4 - 1 \big]-\big[ 2 - 4 - 8 \big]  & 62.5 & 78.1 \\
\midrule
\big[16 - 8 - 4 - 2 \big]-\big[ 1 - 2 - 4 \big]  & 77.1 & 78.1 \\
\big[16 - 8 - 4 - 2 \big]-\big[ 1 - 2 - 8 \big]  & 70.7 & 78.1 \\
\big[16 - 8 - 4 - 2 \big]-\big[ 1 - 4 - 8 \big]  & 64.2 & 78.1 \\
\big[16 - 8 - 4 - 2 \big]-\big[ 2 - 4 - 8 \big]  & 59.1 & 78.1 \\
\bottomrule[1.2pt]
\end{tabular}}
\caption{Comparison of the computations and performance of our model with different reduction ratio at inference. During training, \big[Encoder\big]-\big[Decoder\big] reduction ratio was \big[8, 4, 2, 1\big]-\big[1, 2, 4\big], and it is notated with $^{\dagger}$. GFLOPs were computed with the input size of $2048\times1024$. The optimal inference reduction ratio is in \colorbox{Gray}{gray}.}
\label{tab:tableA}
\end{table}

\section{Various Reduction Ratios for ISR}
\label{C}
In \Cref{tab:tableA}, we compared computational cost (FLOPs) and mIoU performance of our EDAFormer-T under various ISR conditions (\textit{i.e.} the reduction ratio) on Cityscapes dataset. We experimented with 20 number of conditions by increasing the reduction ratio of each encoder stage and decoder stage to demonstrate more results of various reduction ratio than the Table 4 of the main paper. Firstly, the results showed that there is no mIoU performance degradation when the reduction ratio was increased in each decoder stage of our all-attention decoder. These results indicate that the decoder composed of the embedding-free attention is effective to apply our ISR. Secondly, the mIoU performance was maintained when the reduction ratio increased in both 1st and 2nd encoder stages. However, the performance decreased by 0.6\% (\textit{i.e.} 78.7\% $\to$ 78.1\%) when the reduction ratio increased in both the 3rd and 4th encoder stages. Therefore, we suggest the reduction ratio of \big[16,8,2,1\big]-\big[2,4,8\big] as the optimal condition of our ISR, but the user can selectively leverage other conditions if lower computational cost is required even with the performance degradation. 


\section{FPS Comparison with Other Segmentation Models}\label{fps}
In Table \ref{tab:various}, we present the inference speed comparison without any additional accelerating techniques. For fair comparison, we measured Frames Per Second (FPS) of a whole single image of 2048$\times$1024 on Cityscapes using a single RTX3090 GPU. Compared to previous segmentation methods, our method achieved a FPS improvement with a higher mIoU score. 

\begin{table}[t]\centering 
\renewcommand{\arraystretch}{}
\resizebox{0.7\columnwidth}{!}
{
\small
\begin{tabular}{L{4cm}|C{2cm}|C{2cm}|C{2cm}|C{2.5cm}}
\toprule
Model & {Params (M)} & GFLOPs $\downarrow$&{mIoU (\%) $\uparrow$} & FPS (img/s) $\uparrow$ \\ 
\midrule

SegFormer-B0 & 3.8 &125.5& 76.2 & 8.8 \\
FeedFormer & 4.5 & 107.4&77.9 & 9.3 \\
VWFormer-B0 & 3.7 & -&77.2 & 8.9 \\

\midrule
\cellcolor{Gray}EDAFormer-T (w/\, ISR)  &\cellcolor{Gray}4.9 &\cellcolor{Gray}94.9&\cellcolor{Gray}78.7 &\cellcolor{Gray}\textbf{13.3} \\
\bottomrule
\end{tabular}
}\vspace{0.1cm}
\caption{FPS comparison with other segmentation models on Cityscapes. FPS is tested on a single RTX 3090 GPU. }
\label{tab:various}
\end{table}

\section{Applying EFA to Other Backbones}\label{backbones}
For in-depth analysis, we analysed the effectiveness of our embedding-free structure on other backbones in Table \ref{tab:parameter}. In each backbone, we applied our method and added the number of the attention blocks for fair model size. Compared to the other two methods, our method showed 1.5\% and 0.8\% higher accuracy with similar parameter size and the same computational cost. These results demonstrated that our method is also effective for the other transformer-based encoders.

\begin{table}[t]\centering 
\renewcommand{\arraystretch}{}
\resizebox{0.7\columnwidth}{!}
{
\small
\begin{tabular}{C{2cm}|C{3cm}|C{2cm}|C{2cm}|C{3cm}}
\toprule
Model & Method &  {Params (M)} & {MFLOPs $\downarrow$} & {Top-1 Acc. (\%)} \\ 
\midrule
\multirow{2}{*}{PVT} &{original} &  13.2 & 1.9 & 75.1\\
& \cellcolor{Gray}{w/o embedding} & \cellcolor{Gray}13.3 & \cellcolor{Gray}1.9 & \cellcolor{Gray}\textbf{76.6}\\
\midrule
\multirow{2}{*}{PVT v2}& {original} & 3.7 & 0.6 & 70.5 \\
& \cellcolor{Gray}{w/o embedding}  &   \cellcolor{Gray}3.6 & \cellcolor{Gray}0.6 & \cellcolor{Gray}\textbf{71.3}\\
\bottomrule
\end{tabular}
}\vspace{0.1cm}
\caption{Applying our EFA to other backbones on ImageNet-1K.}
\label{tab:parameter}
\end{table}

\begin{figure}[t]
     \centering
     \includegraphics[width=0.97\textwidth]{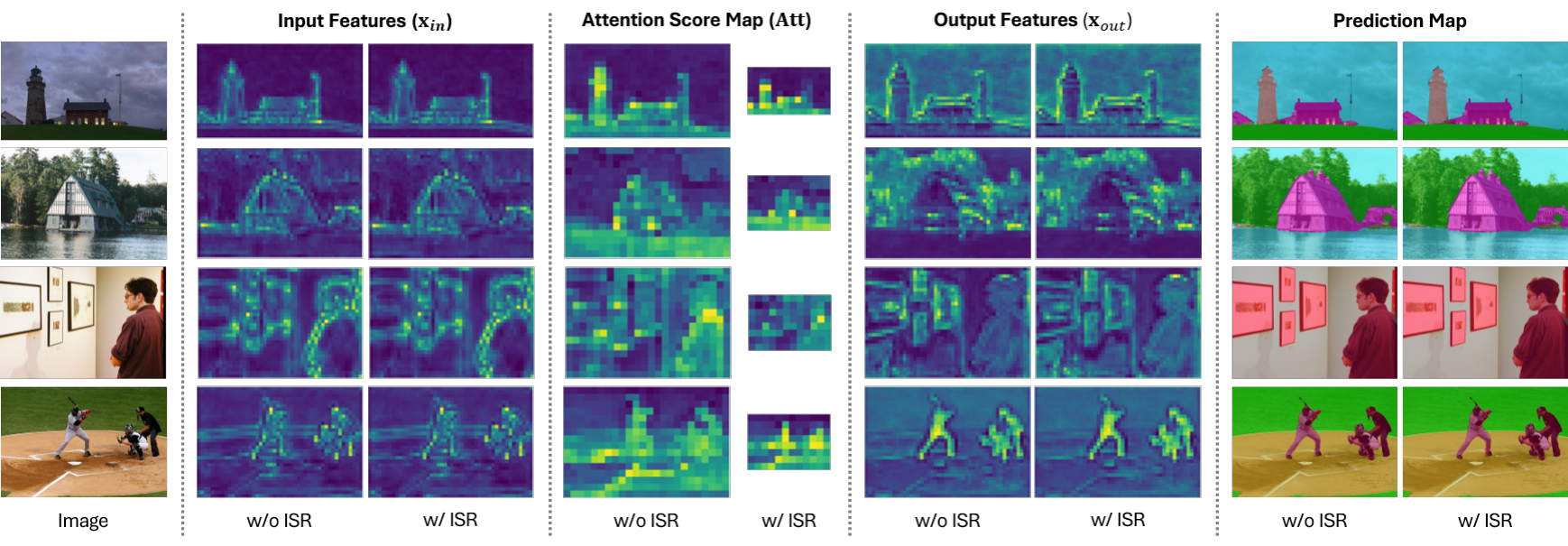} \vspace{-0.3cm}
     \caption{Visualizing the input features of the self-attention, the attention score maps, the output features of the attention, and the prediction maps. `w/o ISR' represents our EDAFormer-T with the base reduction ratio of [8,4,2,1]-[1,2,4]. `w/ ISR' represents our EDAFormer-T applied our ISR method with the reduction ratio of [16,8,2,1]-[2,4,8]. } 
     \label{fig:feat}
\end{figure}

\begin{figure}[t]
     \centering 
     \includegraphics[width=0.99\textwidth]{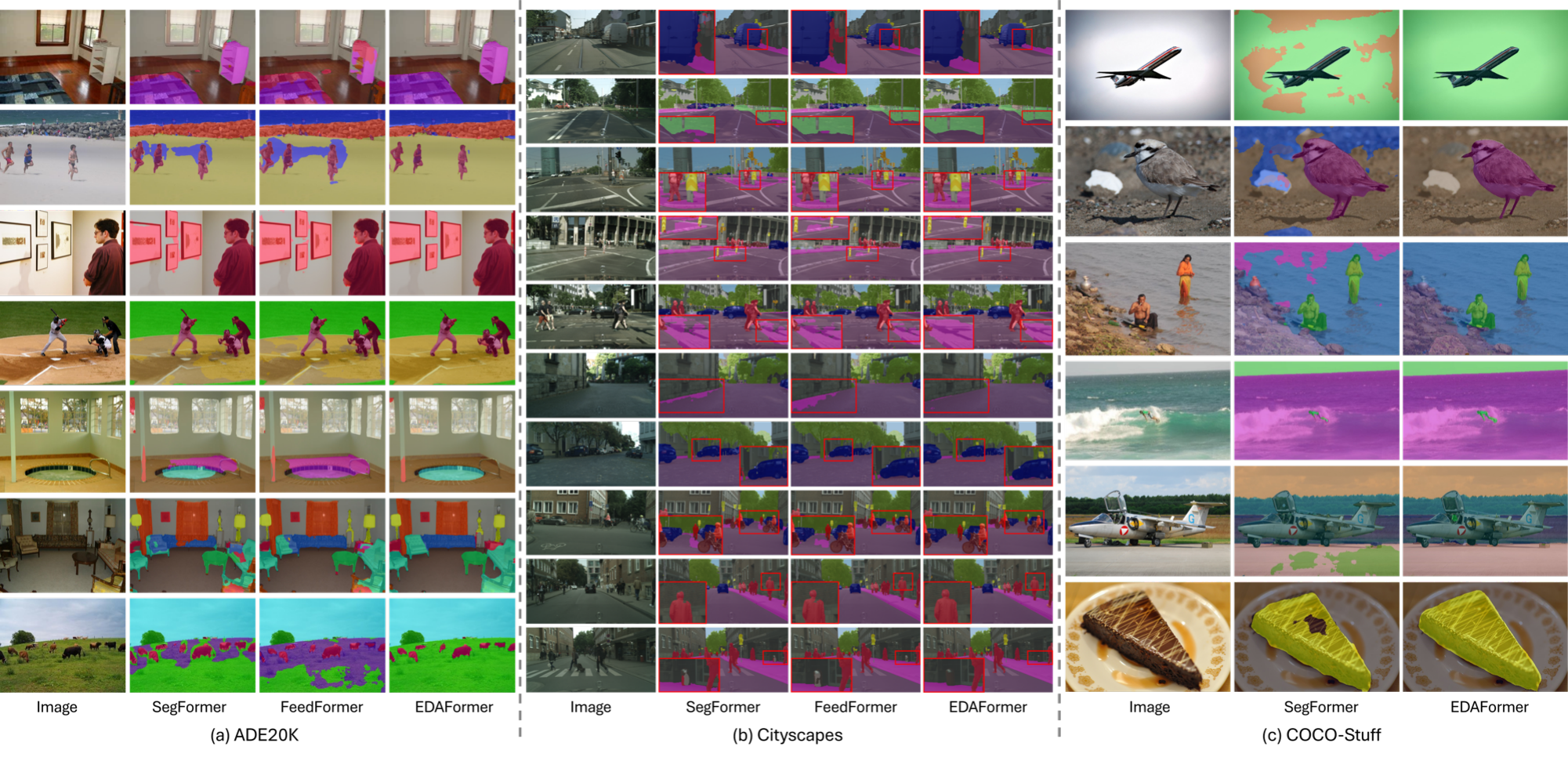} \vspace{-0.4cm}
     \caption{Visualization of qualitative results on ADE20K, Cityscapes and COCO-Stuff. Compared to previous state-of-the-art semantic segmentation models (\textit{i.e.}, SegFormer and FeedFormer), our EDAFormer predicts more precisely for various categories.} 
     \label{fig:qualisup}
\end{figure}

\section{Additional Feature Visualization}
\label{D}
In Fig. \ref{fig:feat}, we visualized the input-output features of our embedding-free attention, the attention score maps, and the predictions before and after applying ISR. The attention regions in the attention score maps applying our ISR were well maintained in comparison to those without ISR, even though the key and value tokens were reduced. In addition, compared to without ISR, the output features with ISR preserved the spatial information by leveraging the self-attention mechanism where the number of the key-value tokens does not affect the input-output spatial structure. As a result, the prediction maps with our ISR were also largely identical to those without ISR.

\section{Additional Qualitative Results}
\label{E}
Qualitative results of our EDAFormer and other state-of-the-art models were illustrated in Fig. \ref{fig:qualisup} on ADE20K \cite{zhou2017scene}, Cityscapes \cite{cordts2016cityscapes} and COCO-Stuff \cite{caesar2018coco}. SegFormer \cite{xie2021segformer} and FeedFormer \cite{shim2023feedformer} were comparatively analyzed for AED20K and Cityscapes, while COCO-Stuff was exclusively compared with SegFormer. Compared to previous methods, our EDAFormer not only presented better performance for large regions, but also exhibited more precise and detailed predictions for boundary regions. These results demonstrate that our EDAFormer, an encoder-decoder attention structure based on EFA, is an efficient yet powerful network for semantic segmentation.

\end{document}